\documentclass[10pt,journal,compsoc]{IEEEtran}

\ifCLASSOPTIONcompsoc
  \usepackage[nocompress]{cite}
\else
  \usepackage{cite}
\fi

\usepackage{amsmath,amsfonts}
\usepackage{array}
\usepackage{textcomp}
\usepackage{stfloats}
\usepackage{url}
\usepackage{verbatim}
\usepackage{graphicx}
\usepackage{multirow}
\usepackage{graphicx}
\usepackage{subfloat}
\usepackage{subfigure}
\usepackage[mathscr]{eucal}
\usepackage{multicol}
\usepackage{algorithm}
\usepackage{algorithmicx}
\usepackage{algpseudocode}
\usepackage{pifont}
\usepackage{float}
\usepackage{multirow}
\usepackage{booktabs}
\usepackage[table]{xcolor}
\usepackage{colortbl}
\usepackage{bm}
\usepackage{utfsym}

\usepackage{enumitem}

\definecolor{gray}{gray}{0.5}

\usepackage{cite}
\usepackage[colorlinks,
            linkcolor=black,
            anchorcolor=black,
            urlcolor=black,
            citecolor=black]{hyperref}

\begin{document}

\title{A Collaborative Multi-Modality Interaction \\for VLA-based End-to-End Autonomous Driving}

\author{Jingtao~Sun,
        Xiaohai~He,
        Yike~Zhang,
        Dong~Huang,
        Yaonan Wang,\\
        Ajmal Mian,~\IEEEmembership{Senior Member,~IEEE}
        and~Mike Zheng Shou,~\IEEEmembership{Senior Member,~IEEE}% <-this % stops a space
\IEEEcompsocitemizethanks{\IEEEcompsocthanksitem  
Jingtao Sun, Xiaohai He, Dong Huang and Mike Zheng Shou are with Department of Electrical and Computer Engineering and Show Lab, National University of Singapore (NUS), Singapore 117583.

\IEEEcompsocthanksitem Yike Zhang, Yaonan Wang are with the School of Artificial Intelligence and Robotics and National Engineering Research Centre for Robot Visual Perception and Control, Hunan University, Changsha 410082, China.

\IEEEcompsocthanksitem Ajmal Mian is with Computer Science and Software Engineering, The University of Western Australia (UWA), Australia, 2006.
}% <-this % stops an unwanted space

\thanks{Received xx xxx 2026; accepted xx xxx 2026. Date of publication
xx xxx 2026; date of current version xx xxx 2026. This research is supported by the National Research Foundation, Singapore under its AI Singapore Programme (AISG Award No: AISG3-RP-2022-030).
}

\thanks{Jingtao Sun and Xiaohai He contributed equally to this work.}
\thanks{Corresponding author: Mike Zheng Shou (mike.zheng.shou@gmail.com).}
\thanks{Project page is available at : \url{https://github.com/S-JingTao/CMMI}.}}

\markboth{Journal of \LaTeX\ Class Files,~Vol.~14, No.~8, August~2015}%
{Shell \MakeLowercase{\textit{et al.}}: Bare Demo of IEEEtran.cls for Computer Society Journals}

\IEEEtitleabstractindextext{%
\begin{abstract}
Vision-Language-Action (VLA) models have emerged as a powerful paradigm for end-to-end autonomous driving by jointly integrating perception, reasoning, and decision making within a unified multimodal framework. However, most existing VLA models formulate end-to-end autonomous driving as a visual question answering task, leading to unreliable and less interpretable decision reasoning. In addition, they fail to establish effective multi-modal interaction across heterogeneous sensors, thereby limiting robust scene perception and reliable driving reasoning in long-tail driving scenarios. To this end, we propose a robust VLA-based end-to-end autonomous driving system that combines multi-modality interaction with  multi-trajectory planning and optimization, enabling more reliable, interpretable, and safer driving decisions. Our method comprises three core components: (1) Affinity-Guided Optimal Transport for main-auxiliary modality two-way interaction; (2) Distribution-Consistent Modality Transfer for heterogeneous modality distribution transfer and cross-modal interaction; (3) Multi-modal Multi-Trajectory Planning along with Perception-Oriented Trajectory Refinement for better driving decisions to long-tail driving scenarios. Experimental results in open-loop and closed-loop datasets demonstrate improvements in safety long-horizon driving reasoning and road scene perception over existing driving systems, highlighting the ability of our mutli-modality interaction and multi-trajectory planning and optimization for scalable VLA-based systems.
\end{abstract}
\begin{IEEEkeywords}
End-to-end autonomous driving, multi-modality fusion, optimal transport, vision-language-action models.
\end{IEEEkeywords}}

\maketitle

\IEEEdisplaynontitleabstractindextext

\IEEEpeerreviewmaketitle

\IEEEraisesectionheading{\section{Introduction}\label{sec:introduction}}

\IEEEPARstart{E}{nd}-to-end autonomous driving (E2E-AD) has gained significant attention in recent years~\cite{chen2024end}, benefiting from advances in multi-modal scene perception, such as object detection~\cite{song2024robustness}, semantic segmentation~\cite{li2023mseg3d}, tracking~\cite{lin2024echotrack}, and online mapping~\cite{zhang2023online}, which provide rich environmental understanding for learning driving policies directly from raw sensor inputs: Camera, LiDAR and IMU \emph{et al}. Unlike traditional modular autonomous driving pipelines, which perform perception, prediction, and planning in a sequential manner, which may result in unsafe or inconsistent driving behaviors, E2E-AD offers several advantages. It unifies perception, prediction, and planning into a single model that can be trained jointly, enabling the entire system to be directly optimized for the final driving task, and its data-driven nature allows performance to scale with more training data and larger models.

Recently, vision-language-models (VLMs) and vision-language-actions (VLAs) are well known for their extensive knowledge and strong reasoning ability, which have been explored for autonomous driving~\cite{li2025spacedrive}, either as assistants~\cite{wang2025omnidrive,shao2024lmdrive} to or replacements~\cite{xu2024drivegpt4} for traditional E2E-AD systems. By expressing driving tasks in natural language~\cite{sima2024drivelm}, these methods enable flexible scene understanding, motion prediction based on semantic cues. E2E driving task also be formulated as a question-answering problem~\cite{wang2026learning,nie2024reason2drive} and construct corresponding benchmarks~\cite{qian2024nuscenes}. However, most VLMs are probabilistic models trained on large-scale 2D data, which inherently constrains their ability to generalize from 2D understanding and reasoning to 3D geometric and spatial representations. This limitation leads to unreliable perception and reasoning in complex 3D driving environments. Given that autonomous driving is a safety-critical system, such limitations hinder the development of interpretable and controllable decision-making mechanisms for E2E-AD. Besides, VLA-based AD typically involves multi-modal alignment and fusion, and there are conflicts in information across different modalities. Early fusion~\cite{zhou2019does,xiao2020multimodal} and late fusion methods~\cite{sobh2018end,chitta2022transfuser} lack structured semantics and their fusion strategies are often heuristic, and lack detailed theoretical foundation to solve this weakness. A fundamental challenge arises from the fact that different modalities should produce physically consistent observations of the same driving scene, while current VLM/VLA-based models lack explicit mechanisms to enforce such cross-modal consistency.

In this regard, we propose to develop a causality-guided multimodal interaction framework for VLA-based end-to-end autonomous driving, which unifies scene perception and motion planning. The framework jointly leverages egocentric camera and LiDAR inputs along with natural-language instructions to produce both causal reasoning and accurate ego-vehicle motion planning, while satisfying stringent efficiency and safety requirements. However, realizing this unified, causality-driven framework remains some challenging due to heterogeneous multimodal inputs, limited 3D understanding and reasoning in current VLMs or VLAs, and the lack of explicit safety-aware decision mechanisms, leading to the following key gaps:

\begin{figure*}[t]
    % \vspace{-0.9cm}
    \centering
    \includegraphics[width=\textwidth]{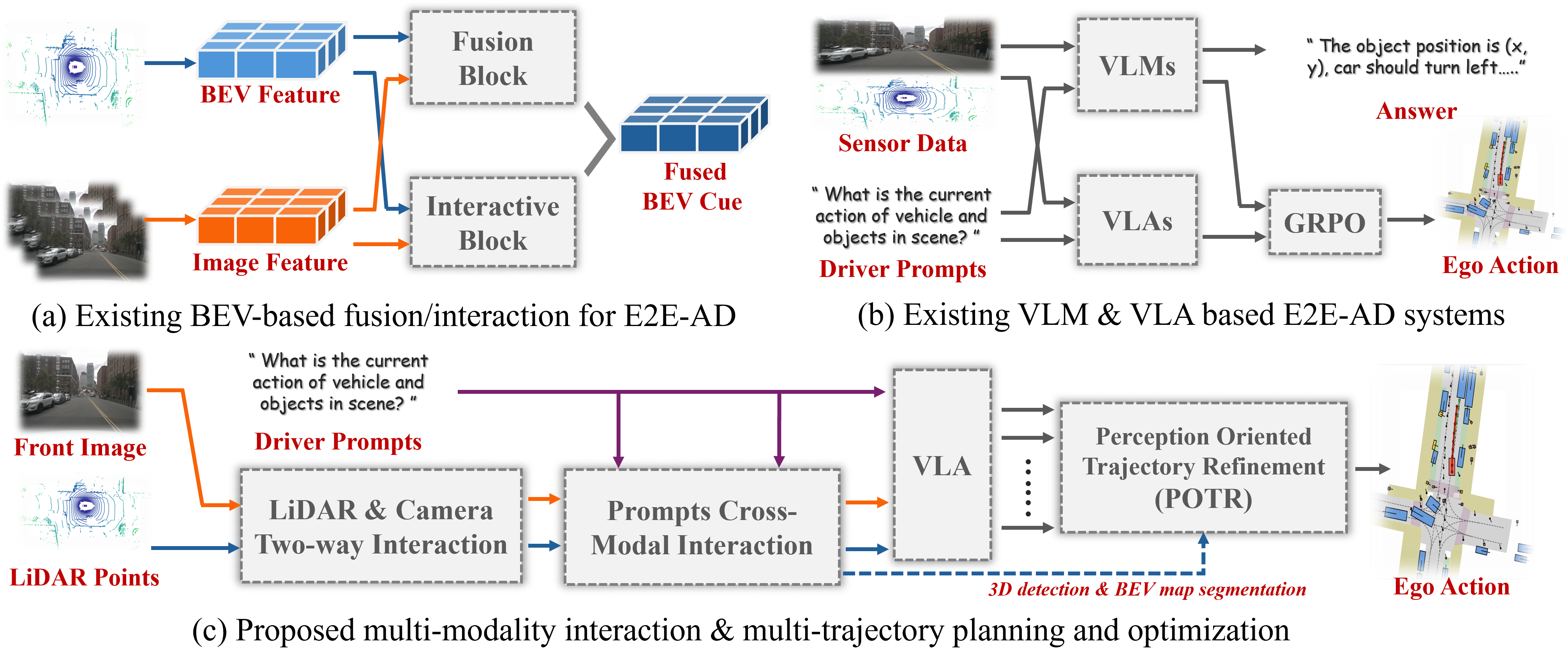} 
    \vspace{-0.9cm}
    \caption{Comparison of different E2E-AD systems. (a)~Existing fusion \& interaction methods for E2E-AD normally fuse individual per-modality into a single hybrid BEV cue. (b) VLM or VLA model focuses more on answer question manner, they lack interpretability and reliable control decisions. (c) Our method achieves multi-modality interaction via optimal transport and generates optimized tarjectory, enhancing the driver trust in E2E-AD.}
    \label{FIG_comparison_method}
    \vspace{-0.4cm}
\end{figure*}

\begin{enumerate}[label=\arabic*)] 
  \item \textbf{Cross-modality misalignment \& heterogeneity.} The fundamental challenge is to construct consistent and reliable scene embeddings from different, misaligned, and imperfect multimodal inputs (\emph{e.g.}, RGB images and LiDAR point clouds). Moreover, incorporating natural language as fine-grained knowledge and high-level instructions for controlling visuomotor agents in VLM-based autonomous driving, introduces additional complexity, requiring more effective modal information interaction and complementary integration.
  \item \textbf{Inherent gaps of VLAs for 3D streetscape perception.} Current VLMs/VLAs are limited in handling 3D tasks such as geometric measurement and distance estimation, which are critical for autonomous driving. There are two reasons: the lack of 3D pre-training, leading to poor alignment between 3D coordinates and object semantics and yielding weak inference of inter-object spatial relations; The digit-by-digit processing of numerical
symbols in VLMs/VLAs restricts accurate reasoning, thereby degrading precision in continuous waypoint predictions.
  \item \textbf{Safety-critical explicit reasoning \& decision making.} Existing E2E-AD systems typically rely on implicit decision making, without explicitly modeling safety-related considerations such as online interactions with surrounding vehicles/pedestrians and evolving risks (collision, lane departure). This limits their ability to make reliable decisions and handle critical situations in real-time long-tail driving conditions.
\end{enumerate} 
%Moreover, most VLMs are trained on internet data, lacking spatial understanding and specialized training in the field of AD, making it difficult for them to fully adapt to dynamic and complex driving scenarios.

To bridge these gaps, we introduce an interpretable VLA-based E2E-AD system with reliable driving decisions, leveraging a universal multi-modality interaction strategy, as depicted in Fig.~\ref{FIG_comparison_method}(c). First, we propose multi-modality interaction using discrete optimal transport plan for modal-specific learning, addressing a foundamental limitation of the previous fusion/interaction methods depicted in Fig.~\ref{FIG_comparison_method}(a). Concretely, we formulate LiDAR and camera interactive fusion by employing an Affinity-Guided Optimal Transport that builds similarity measure between main and auxiliary modality to achieve two-way transport plan and interaction. Furthermore, we achieve prompts corss-modal interaction by integrating heterogeneous features into a unified mainfold, by implementing a Distribution-Consistent Modality Transfer that transforms heterogeneous modal distributions into a unified Gaussian space.

Additionally, we present the Multi-Modality Multi-Trajectory Planning strategy along with Perception-Oriented Trajectory Refinement, constructing a driving risk cost map using real-time 3D
detection and BEV map segmentation, significantly enhancing long-tail scene driving capacity and deep reasoning abilities. The proposed optimization method directly refines generated trajectories to improve the VLA model's driving reliability without additional policy training, unlike Group Relative Policy Optimization (GRPO), that needs to optimize policy parameters through reward-driven updates, as depicted in Fig.~\ref{FIG_comparison_method}(b). Overall, our contributions are summarized as follows:

\begin{itemize}
\item We propose the Affinity-Guided Optimal Transport for main-auxiliary modality two-way interaction for autonomous driving tasks, by adaptively learning the interaction relationship of different pairwise modalities along with affinity calculation.

\item We introduce the Distribution-Consistent Modality Transfer for heterogeneous multi-modality distribution transfer and interaction, by designing semantic-aligned normalizing flow, which ensures modal distribution consistency for cross-modal interaction.

\item We present the Multi-modal Multi-Trajectory Planning along with Perception-Oriented
Trajectory Refinement, which exploit the driving risk cost map to optimize dirving decisions, leading to better generalization to long-tail driving scenarios.

\end{itemize}

\newpage
\section{Related Work}

\subsection{Multi-Modality Fusion for Autonomous Driving}
Multi-modality fusion~\cite{fan2025mgaf} for autonomous driving aims to complement RGB images with depth or 3D semantics to improve driving performance. Early explorations~\cite{zhou2019does,xiao2020multimodal} demonstrate the effectiveness of image semantics and point cloud 3D space information as explicit intermediate representations for driving. Another concurrent works perform relatively straightforward fusion strategies, such as late fusion~\cite{sobh2018end} or feature-level augmentation~\cite{natan2022end,chen2022learning} to integrate semantic and geometric cues. However, they often rely on loosely coupled fusion mechanisms~\cite{vora2020pointpainting}, which can limit cross-modal interaction and lead to suboptimal performance in more complex end-to-end driving scenarios.
Existing multi-modal fusion strategies in E2E autonomous driving can be broadly divided into two
categories: flatten fusion and BEV-based fusion. Flattened fusion methods~\cite{chitta2022transfuser,shao2023safety,chen2022autoalign} usually project features from images and LiDAR point clouds into a common latent space, where interactions are modeled using attention mechanisms. These methods are attractive because they are flexible, computationally efficient, and often do not rely heavily on precise geometric calibration. However, since they do not explicitly preserve 3D spatial structure, they lack strong spatial interpretability and will tend to perform worse outcome in tasks that demand accurate spatial reasoning. In contrast, BEV fusion methods~\cite{jia2023think,li2024bevformer,ye2023fusionad} first project multi-modal features into a unified bird’s-eye view coordinate system, enabling geometric alignment across different sensors. 

In this work, we propose a multi-modal interaction that differs from previous fusion-based approaches. Our key insight is to achieve cross-modal interaction via semantic-aware transport plans, where we exchange same semantic contents while preserving modality-specific information, thereby achieving more effective cross-modal interaction.

\subsection{End-to-End Autonomous Driving}
Traditional end-to-end autonomous driving~\cite{wu2022trajectory,zeng2019end} aims to directly predict the vehicle path and low-level control signals only based on visual inputs. Recent advances in E2E-AD~\cite{li2024bevformer,zhang2024seflow} have evolved from traditional modular stacks to fully differentiable frameworks with an emphasis on motion planning. UniAD~\cite{hu2023planning} represents an early attempt toward unified end-to-end autonomous driving, jointly modeling multiple perception tasks to enhance planning performance. VAD~\cite{jiang2023vad} introduces compact vectorized scene representations to improve efficiency, that adopt single-trajectory planning for enhanced performance. A line of E2E-AD works~\cite{chitta2022transfuser,gu2024producing,li2024ego} adopt a single-trajectory planning paradigm to improve performance. However, such approaches often suffer from limited action diversity, which may introduce potential safety risks. While VADv2~\cite{jiang2024vadv2} further extends this line by enabling multi-modal planning through scoring and sampling from a predefined set of anchor trajectories. Existing multi-mode trajectory planning methods, such as SparseDrive~\cite{sun2025sparsedrive}, MomAD~\cite{song2025don} and DiffusionDrive~\cite{liao2025diffusiondrive}, remain limited in addressing mode collapse in imitation learning. Thereby, DriveSuprim~\cite{yao2026drivesuprim} improves selection-based multi-modal trajectory generation via a coarse-to-fine candidate refinement strategy combined with rotation augmentation and self-distillation, yielding strong safety and trajectory quality.

However, these E2E-AD systems mainly learn an implicit mapping from sensory observations to control commands, limiting their ability to interact with humans through natural language instructions. In contrast, our method integrates multi-modal interaction with trajectory planning and optimization, enabling effective human-vehicle interaction and more reliable driving decisions.

\subsection{VLM \& VLA in Autonomous Driving Agents}
Compared to traditional E2E designs, VLM/VLA-based E2E models can offer the potential for enhanced generalization, allowing them to better cope with complex and dynamically evolving driving environments. VLMs or VLAs for end-to-end autonomous driving~\cite{li2026spacedrive,wang2026learning,shao2024lmdrive,tian2024drivevlm,xu2026wam,xu2024drivegpt4} aim to map egocentric driving-view video inputs and natural-language instructions into both causal reasoning and precise ego-vehicle motion planning, while satisfying stringent efficiency and safety requirements. Early works, such as DriveGPT4~\cite{xu2024drivegpt4}, VLP~\cite{pan2024vlp} and DriveVLM~\cite{tian2024drivevlm}, take E2E driving as a language-conditioned sequence task, leveraging large vision-language models for low-level ego controls and street scene understanding.
Existing methods can be broadly divided into two categories: dual-system and unified paradigms. Dual-system approaches~\cite{hou2025driveagent,zhang2025epona,zhang2025adadrive} incorporate autoregressive VLMs~\cite{liu2023visual,touvron2023llama} as auxiliary reasoning modules to generate high-level semantic guidance, which is subsequently consumed by dedicated motion planners, that are typically based on diffusion-driven iterative optimization~\cite{feng2025artemis,ho2020denoising,jiang2025diffvla,liao2025diffusiondrive} to produce feasible trajectories. Furthermore, unified approaches~\cite{zeng2026futuresightdrive,zhou2026autovla,hwang2024emma,chen2025drivinggpt}, such as EMMA~\cite{hwang2024emma} and DrivingGPT~\cite{chen2025drivinggpt}, directly cast planning as a language generation problem, allowing reasoning and action prediction to be performed jointly within a single VLM framework. 

However, these methods largely formulate E2E-AD as a visual question answering task, where driving decisions are generated by answering language queries, lacking explicit mechanisms for making reliable motion decision. Our approach addresses these limitations through multi-modal interaction and online trajectory optimization.

\subsection{Optimal Transport}
Optimal transport is a well-established field of mathematics founded by Gaspard Monge
and Leonid Kantorovich~\cite{kantorovich2006translocation}. It has offered new solutions to different problems in machine learning, such as generative modeling and
transfer learning. There are mainly three types of generative models that
benefit from optimal transport, namely, GANs~\cite{goodfellow2020generative}, Variational
Autoencoders~\cite{pinheiro2021variational} and normalizing flows~\cite{kobyzev2020normalizing}. Transfer learning~\cite{pan2009survey} aims to transfer knowledge across domains whose data are drawn from different probability distributions. Meanhwile, optimal transport in this area has shown better performance in various fields, such as image classification~\cite{courty2016optimal}, sentiment analysis and fault
diagnosis~\cite{montesuma2021wasserstein}. In this paper, we apply discrete optimal transport to achieve multi-modality interaction for end-to-end autonomous driving, leveraging its valuable ability of transferring knowledge between different domains to support interaction among heterogeneous modalities.

% \subsection{Flow Matching}
% Flow Matching~\cite{lipman2022flow} provides a different paradigm for generative modeling, that learns a continuous vector field that directly transports samples from a simple prior distribution to the data distribution.
% Despite flow matching has achieved the remarkable success in the generating coutinuous spatial signals, such as images~\cite{esser2024scaling,wang2026fudoki} and video~\cite{wu2023tune}, their performance still falters when applied to discrete sequential data. Recent studies~\cite{gat2024discrete,shaul2025flow} have also extended flow matching to discrete data domains,  which extend diffusion processes to tokenized sequences. WAM-Flow~\cite{xu2025wam} has presented a straightforward application of discrete flow matching to VLA model for autonomous driving. 
% In this paper, we apply discrete flow matching
% to VLM for end-to-end autonomous driving, leveraging its inherent parallel generation capability to support multi-modal multi-trajectory planning and optimization.

\newpage
\section{Preliminaries}

\subsection{Problem \& Task Definition}
Different from traditional E2E-AD systems, such as~\cite{kim2026safedrive,wang2026learning,chitta2022transfuser,liao2025diffusiondrive}, that only aim to learn expert driving
policies by mapping raw sensor inputs to ego-vehicle trajectory
outputs, we hope to extend the proposed approach into an end-to-end multi-task learning. To demonstrate the universality and versatility of our multi-modal interaction strategy, we simultaneously resolve road scene perception and ego-planning tasks, including auxiliary tasks: 3D object detection and BEV map segmentation, along with primary task: motion prediction, that is formulated as a multi-input-multi-output (MIMO) structure.

Given a front-view image $I \in {\mathbb R^{3 \times H \times W}}$ and LiDAR point cloud ${\rm P} \in {\mathbb R^{n \times 3}}$ with their transformation matrices, we adopt a two-stage task objective. In the first stage, the multi-modal interactive model is trained for 3D object detection and BEV map segmentation, while producing fused camera visual token and LiDAR BEV token. In the second stage, the pretrained model is fine-tuned for trajectory prediction, aiming to generate the ego vehicle motion trajectory $\tau  \in {\mathbb R^{{T_f} \times 2}}$ under guidance of language navigation prompts. 
\begin{align}
\tau  = \{ ({x_t},{y_t})\} _{t = 1}^{{T_f}},~~({x_t},{y_t}) \in {\mathbb R^2},
\end{align}
where $(x_t, y_t)$ and $T_f=8$ denote the 2D waypoint coordinate at time t and the number of planning horizon. 
%Motion trajectory should conform to the road topology, avoid static and dynamic obstacles, and reflect plausible high-level driving behaviors (e.g., turning, yielding, or going straight).

\subsection{Background: Discrete Optimal Transport}
Optimal Transport~(OT)~\cite{montesuma2024recent} provides a principled search plan for measuring the discrepancy between probability distributions by explicitly modeling the transportation of probability mass, which estimates a globally optimal correaltion between two distributions while minimizing the total transportation cost $c$. Consider two probability measures $\mu  \in \mathcal P(\mathcal X)$ and~$\nu  \in \mathcal P(\mathcal Y)$, where $\mathcal P( \cdot )$ denotes the space of probability measures. In
its Kantorovich Relaxation formulation, OT estimates a joint probability distribution $\gamma (x,y)$ and the optimization problem becomes as follows:
\begin{align}
\mathop {\arg \min }\limits_{\gamma  \in \Pi (\mu ,\nu )} \int\limits_{\mathcal X \times \mathcal Y} {c(x,y)d\gamma (x,y)} ,
\end{align}
where ${\Pi (\mu ,\nu )}$ denotes the set of all transport plans.

Likewise, for Discrete Optimal Transport, its probability distribution is represented by finite feature sets. Suppose two discrete token sets are $\mathcal X = \{ {x_i}\} _{i = 1}^N$, $\mathcal Y = \{ {y_j}\} _{j = 1}^M$, and the transportation cost matrix is defined as ${\mathbf C_{ij}} = c({x_i},{y_j})$, along with the transport plan matrix is represented by $\mathbf P \in \mathbb R_ + ^{N \times M}$, the entropy regularization of discrete OT objective can be expressed as:
\begin{align}\label{transport_plan}
\mathop {\arg \min }\limits_{\mathbf P \in R_ + ^{N \times M}} \left\langle {\mathbf P,\mathbf C} \right\rangle  - \varepsilon H(\mathbf P),
\end{align}
\begin{align}\label{transport_plan_more}
H(\mathbf P) =  - \sum\limits_{ij} {{\mathbf P_{ij}}\log {\mathbf P_{ij}}} ,
\end{align}
where $\left\langle { \cdot , \cdot } \right\rangle $ is the Frobenius inner product and $H(\mathbf P)$ is the entropy of the transport plan. 
The transport matrix satisfies: $\mathbf P \cdot {\rm{\vec 1 = }}\mu ,~{\mathbf P^T} \cdot {\rm{\vec 1 = }}\nu ,~\mathbf P \ge 0$, and ${{\rm{\vec 1}}}$ is a vector of ones.
Entropy regularization smooths the transport matrix while transforming discrete OT into a strictly convex optimization problem. The regularized discrete OT problem admits the following closed-form solution:
\begin{align}\label{plan_solve}
\mathbf P^ * = {\rm{diag}}(\mu ) \cdot K \cdot {\rm{diag}}(\nu )~~\emph{s.t.}~~K = \exp ( - \mathbf C/\varepsilon ) .
\end{align}

\section{Methodology}
\begin{figure*}[t]
    % \vspace{-0.9cm}
    \centering
    \includegraphics[width=\textwidth]{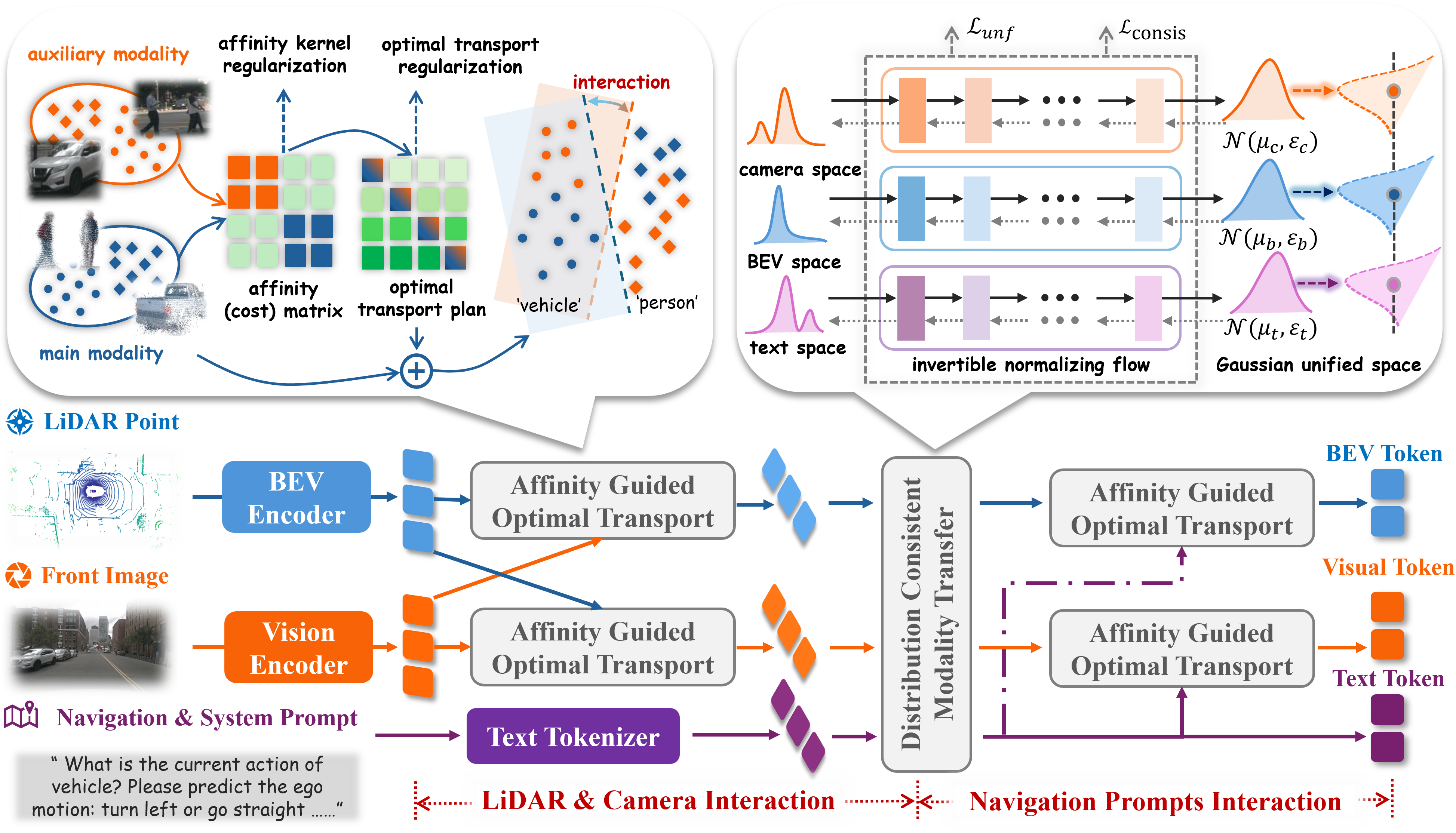} 
    % \vspace{-0.4cm}
    \caption{Architecture of the proposed multi-modality interaction framework. Our method takes as input a front-view image, LiDAR point cloud, and a natural-language navigation command with a system prompt, and outputs the fused visual, BEV and text tokens. "Affinity-guided Optimal Transport " aims to make interactive relations via the similarity computation. "Distribution-Consistent Modality Transfer" is to transform heterogeneous token distributions into a unified latent Gaussian space for subsequent cross-modal interaction. It consists of two fusion processes: LiDAR \& Camera Two-Way Interactive Fusion and Language Navigation Prompts Cross-Modal Interactive Fusion.}
    \label{FIG_overall}
    \vspace{-0.3cm}
\end{figure*}

In this section, we first present our multi-modal interactive paradigm as depicted in Fig.~\ref{FIG_overall}. For each main modality, we compute two-way and cross-modal interactions via Learnable Affinity-guided Optimal Transport (Sec.~\ref{sec4-1}) and Distribution-Consistent Modality Transfer (Sec.~\ref{sec4-2}), while achieving information exchange and aggregation via multi-modal interaction. Then, we introduce how to implement multi-modality multi-trajectory generation~(Sec.~\ref{sec4-3}) and trajectory planning optimization~(Sec.~\ref{POTR}).

\subsection{LiDAR and Camera Interactive Fusion}\label{sec4-1}

Since the similarity computation measures the importance of two modalities, it is feasible to introduce the similarity computation to our multi-modal interaction, meanwhile it can extract various nonlinear relationships and is suitable for processing heterogeneous multi-modal transportation cost matrix. Thus, we first introduce a Learnable \emph{\textbf{Affinity-Guided Optimal Transport}} to enable two-way interaction between main and auxiliary modality. Given the camera visual token ${V} = \{ {v_i} \in {\mathbb R^d}\} _{i = 1}^{{N_v}}$ and the LiDAR BEV token ${B} = \{ {b_i} \in {\mathbb R^d}\} _{i = 1}^{{N_b}}$, let us first take $B$ as the main modality and $V$ as the auxiliary modality as an example to describe how to make interactive relations. The learnable affinity matrix ${\mathbf S_{BV}}$ representing the similarity between the main-modal features and the auxiliary-modal
features, is computed by
\begin{align}\label{sbv}
{\mathbf S_{BV}} = {\rm{MLP}}(\rm{CrossAttn}({V},{B})),
\end{align}
thereby, the transportation cost matrix is generated by the learnable mapping ${\mathbf C_{BV}} = \phi (-{\mathbf S_{BV}})$, which ensures that ${\mathbf C_{BV}}$ is non-negative. On the other hand, the affinity matrix can also be expressed by
\begin{equation}
\resizebox{0.9\linewidth}{!}{$
\mathbf S_{BV}=
\left[
\begin{array}{ccccc}
\kappa(v_1,b_1) & \cdots & \kappa(v_1,b_j) & \cdots & \kappa(v_1,b_{N_b})\\
\vdots & \ddots & \vdots & \ddots & \vdots\\
\kappa(v_i,b_1) & \cdots & \kappa(v_i,b_j) & \cdots & \kappa(v_i,b_{N_b})\\
\vdots & \ddots & \vdots & \ddots & \vdots\\
\kappa(v_{N_v},b_1) & \cdots & \kappa(v_{N_v},b_j) & \cdots & \kappa(v_{N_v},b_{N_b})
\end{array}
\right]
$},
\end{equation}
\begin{align}
\kappa ({v_i},{b_j}) = {({v_i})^T} \cdot {W_{d \times d}} \cdot {b_j},
\end{align}
where ${\kappa ({v_i},{b_j})}$ is a general linear kernel. ${W_{d \times d}}$ is a positive semi-definite matrix and is an adaptive weight to adjust the
multi-modal similarity along with the training stage.

\noindent\textbf{Theorem 1. (Orthogonal Diagonalization)}~\emph{Every real symmetric matrix $W$ can be decomposed as $W = {U^T}\Lambda U$, where $U$ is an orthogonal matrix
and $\Lambda$ is a diagonal matrix with eigenvalues of $W$ as its diagonal elements.}

According to Theorem 1, we can decompose ${W_{d \times d}}$ as an orthogonal
matrix, due to it is a real
symmetric matrix. Thus, the general linear kernel can be defined directly as:
\begin{align}
\kappa ({v_i},{b_j}) = {({v_i})^T} \cdot {U^T}\Lambda U \cdot {b_j} = {(U{v_i})^T} \cdot \Lambda  \cdot (U{b_j}).
\end{align}
In addition, based on Theorem 1, we can also draw another conclusion: for the matrix ${W_{d \times d}}$, it is a positive semidefinite matrix \emph{iff} it is symmetric and all its eigenvalues $\Lambda  = diag({\lambda _1},{\lambda _2}, \ldots ,{\lambda _d})$ are
non-negative. In other words, the matrix ${W_{d \times d}}$ is positive semidefinite if and only if $\Lambda  \ge 0$, \emph{i.e.}, ${\lambda _i} \ge 0,\forall i$, where ${\lambda _i}$ are the eigenvalues of ${W_{d \times d}}$. To encourage ${W_{d \times d}}$ to be a positive semi-definite matrix, so that our model can learn the interactive relations between the main-modal features $B$
and the auxiliary-modal features $V$ better, we propose a affinity kernel regularization loss:
\begin{align}
{\mathcal L_{reg}} = \sum\nolimits_{W \in {\mathbf S_{BV}}} {(|{W^T} - W| + \sum {|{\lambda _i} - |{\lambda _i}||} )} .
\end{align}

To establish a globally optimal modal correspondence, we then formulate the token interaction as a discrete optimal transport problem. Specifically, the transport plan is obtained by minimizing the overall transportation cost under the marginal constraints, according to Eq.~(\ref{transport_plan}). Instead of performing local token matching, the transport plan jointly considers all visual and LiDAR BEV tokens, and we solve this transport plan using the entropy-regularized Sinkhorn algorithm in Eq.~(\ref{plan_solve}). Finally, the transport plan is employed to aggregate cross-modal interaction:
\begin{align}\label{inter_bv}
\tilde V = V \oplus  {\mathbf P^ *} \cdot B,~~\tilde B = B \oplus  {({\mathbf P^ *})^T} \cdot V.
\end{align}
Our
optimal transport regularization loss is defined as follows:
\begin{equation}
\begin{aligned}
{\mathcal L_{ot}} &=\; ||{\mathbf P^ * } \cdot {\rm{\vec 1}} - \mu ||_2^2 + ||{({\mathbf P^ * })^T} \cdot {\rm{\vec 1}} - \nu ||_2^2\\
&+ \sum\limits_{i,j} {\mathbf P_{ij}^ * } ||{v_i} - {b_j}||_2^2 - \sum\limits_{i,j} {\mathbf P_{ij}^ * } \log \mathbf P_{ij}^ *,
\end{aligned}
\end{equation}
where first two items constrains the learned transport plan to satisfy the prescribed marginal distributions.The third item introduces a transport-guided semantic alignment loss to preserve semantic consistency after transport, and last entropy regularization is employed to improve numerical stability during Sinkhorn optimization.

% \newpage
\subsection{Navigation Prompts Interactive Fusion}\label{sec4-2}

The intrinsic heterogeneity among high-level semantic modality of natural-language navigation and low-level observation modality of camera and LiDAR results in inconsistent latent distributions, despites describing the same driving scene. Consequently, modality-specific tokens reside on different feature manifolds, making direct token-level interaction susceptible to semantic misalignment and modality bias. To enable reliable cross-modal interaction, we then employ \emph{\textbf{Distribution-Consistent Modality Transfer}} to transform heterogeneous token distributions into a unified latent Gaussian space to serves as the foundation for cross-modal distribution transfer and feature interaction.

Let $T = \{ {t_i} \in {\mathbb R^d}\} _{i = 1}^{{N_t}}$ denotes navigation command text token, our main idea is to transfer the distributions from observed modalities to semantic one through invertible normalizing flow~\cite{dinh2016density}, and achieve better interaction with high-level distribution consistency. We first use the shared normalizing flow to project camera, LiDAR, and language tokens into a common latent Gaussian mainfold:
\begin{align}
z_i^m = {\mathcal F_\theta }({m_i}) \sim \mathcal N({\mu _m},{\varepsilon _m}), m \in \{ t,c,b\}, 
\end{align}
where ${\mathcal F_\theta }( \cdot )$ denotes proposed normalizing flow model of all modality. This unified normalizing flow is optimized by maximizing the conditional log-likelihood according to the change-of-variable theorem, yielding 
\begin{align}
{\mathcal L_{unf}} =  - \sum\limits_{m \in \{ t,c,b\} } {\sum\limits_i {(\log p(z_i^m) + \log |\det {J_\mathcal F}|)} } ,
\end{align}
where ${p(z_i^m)}$ follows the standard Gaussian distribution $\mathcal N(0,I)$ and ${J_\mathcal F}$ is the Jacobian of invertible transformation. The first term encourages heterogeneous modality tokens to follow a common Gaussian distribution, while the second term preserves the probability measure under the invertible mapping by accounting for the local volume change. However, maximizing the likelihood of each modality independently does not explicitly minimize the distribution discrepancy among different modalities. Therefore, we further introduce cross-modal distribution consistency loss based on the optimal transport to directly reduce the Sinkhorn distance between latent distributions, encouraging camera, LiDAR, and language features to share a common probability measure in the latent space:
\begin{align}
{\mathcal L_{{\rm{consis}}}} = {D_e}({\mathcal P_t},{\mathcal P_c}) + {D_e}({\mathcal P_t},{\mathcal P_b}) + {D_e}({\mathcal P_b},{\mathcal P_c}),
\end{align}
where ${\mathcal P_c},{\mathcal P_d}$ and ${\mathcal P_t}$ denote the transformed distributions of the camera, LiDAR, and language tokens, respectively. ${D_e}( \cdot , \cdot )$ denotes the Sinkhorn distance, that is derived from Eq.~(\ref{transport_plan}) and Eq.~(\ref{transport_plan_more}), and can be expressed as:
\begin{align}
{D_e} = \mathop {\min }\limits_\mathbf P { < \mathbf P,\mathbf C >  - \sum {{\mathbf P_{ij}}\log {\mathbf P_{ij}}} } ,
\end{align}
where $\mathbf P$, $\mathbf C$ denote transport plan and cost matrix. While the unified normalizing flow reduces cross-modal distribution discrepancy, it does not guarantee semantic correspondence between heterogeneous tokens. We therefore apply the proposed Learnable Affinity-guided Optimal Transport to establish semantic-aware token alignment, where semantic information is adaptively transferred to both camera and LiDAR representations:
\begin{align}\label{inter_language}
\hat V = \tilde V \oplus {\mathbf{P}}_{VT}^ *  \cdot T,\;\;\hat B = \tilde B \oplus {\mathbf{P}}_{BT}^ *  \cdot T.
\end{align}
Besides, we apply two task-specific heads to the fused BEV token feature map, which helps us optimize our subsequent tarjectory planning introduced in Sec.~\ref{POTR}. We follow BEVFusion~\cite{liu2023bevfusion} use the same formulation for 3D object detection and map segmentation. Detailed process of our multi-modal interaction is presented in Alg.~\ref{alg:interaction}.

\begin{algorithm}[t]
\caption{Multi-modality Interaction}
\label{alg:interaction}
\begin{algorithmic}[1]

\Require Camera visual token $V$, LiDAR BEV token $B$, language navigation command token $T$
\Ensure Fused visual token $\hat V$ and LiDAR BEV token $\hat B$

\State Compute the affinity matrix $\mathbf S_{BV}$ with Eq.~(\ref{sbv}).
\State Construct the transport cost matrix $\mathbf C_{BV}$.
\State Estimate the best transport plan $\mathbf P^{*}$ via Sinkhorn.
\State Perform camera \& LiDAR interaction with Eq.~(\ref{inter_bv}):
\[\tilde V = V \oplus  {\mathbf P^ *} \cdot B,~~\tilde B = B \oplus  {({\mathbf P^ *})^T} \cdot V\]
\State Project $\tilde V$, $\tilde B$, and $T$ into the unified Gaussian space:
\[z_i^m = {\mathcal F_\theta }({m_i}) \sim N({\mu _m},{\varepsilon _m})\]

\State Compute the affinity matrices $\mathbf S_{VT}$ and $\mathbf S_{BT}$.
\State Estimate the transport plans $\mathbf P_{VT}^{*}$ and $\mathbf P_{BT}^{*}$.
\State Perform language navigation interaction with Eq.~(\ref{inter_language}):
\[\hat V = \tilde V \oplus {\mathbf{P}}_{VT}^ *  \cdot T,\;\;\hat B = \tilde B \oplus {\mathbf{P}}_{BT}^ *  \cdot T\]
\State \textbf{return} $\hat V,\hat B$

\end{algorithmic}
\end{algorithm}

% \newpage
\begin{figure*}[t]
    % \vspace{-0.9cm}
    \centering
    \includegraphics[width=\textwidth]{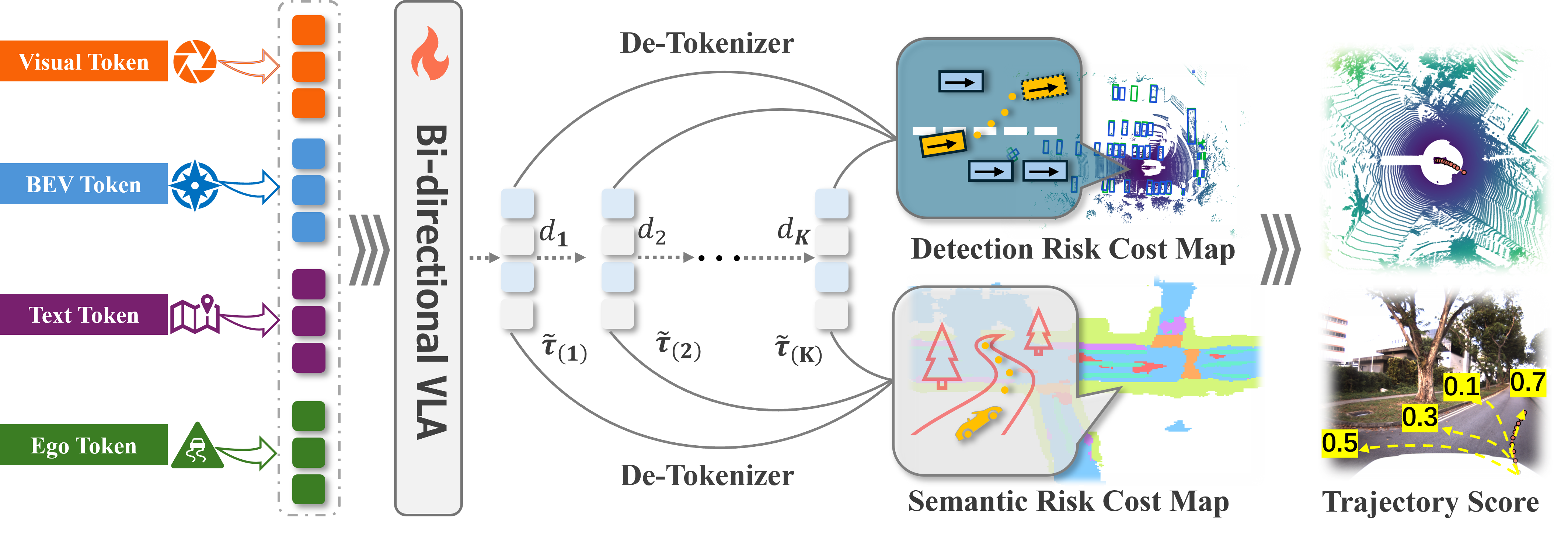} 
    \vspace{-0.5cm}
    \caption{Overview of our multi-modal multi-trajectory planning and trajectory refinement. Given the visual, BEV, Text and vehicle ego-vehicle state tokens, we first generate a sequence of continuous trajectories with an 8-waypoint, and further optimize each trajectory by building a driving risk cost map (including detection risk cost map \& semantic risk cost map), using real-time 3D detection and BEV map segmentation.}
    \label{FIG_plan}
    % \vspace{-0.3cm}
\end{figure*}

\subsection{Multi-Modal Multi-Trajectory Planning}\label{sec4-3}
Inspired by base principle of DFM~\cite{gat2024discrete}, we formulate original trajectory planning task as a conditional multi-modal multi-trajectories generation problem, as shown in Fig. 2. We preserve all intermediate flow states and decode them into candidate trajectories. The denoising evolution naturally forms a trajectory set, enabling multi-modal planning without introducing additional trajectory heads or anchor branches. Taking multimodal discrete token as inputs, including camera visual token, LiDAR BEV token, language navigation token and ego-vehicle state token, we saved the intermediate trajectories after each denoising step and combined them into the final set of trajectories $\mathcal T  = \{ {{\tilde \tau }_{({\rm{1}})}},{{\tilde \tau }_{(2)}}, \ldots ,{{\tilde \tau }_{(K)}}\} $. To respect multi-modal structure of each tokenized space, we present the marginal probability path and define a Gibbs distribution induced by a distance metric $d$. Thus, the trajectory generated in the k-th step is:
\begin{align}
{p_t}({{\tilde \tau }_{(k)}}|{\tau _{{\rm{gt}}}}) = softmax ( - {\beta _t}d({{\tilde \tau }_{(k)}},{\tau _{{\rm{gt}}}})),
\end{align}
where ${\beta _t} \in [0,{\rm{1}}]$ is a monotonically increasing scheduling function and $d({{\tilde \tau }_{(k)}},{\tau _{{\rm{gt}}}})$ is a weighted sum of coordinate-wise dissimilarities. We also consider a Continuous Time Discrete Markov Chain with a transition rate to achieve this marginal probability path and construct a marginalization of conditional probability velocities:
\begin{align}\label{loss_plan}
{u_t}({{\tilde \tau }_{(k)}},{{\tilde \tau }_{(k - {\rm{1}})}}|{\tau _{{\rm{gt}}}}) = {p_t}({{\tilde \tau }_{(k)}}|{\tau _{{\rm{gt}}}}){{\dot \beta }_t}{[{d_{k - {\rm{1}}}} - {d_k}]_ + },
\end{align}
where ${[d]_ + } = \max (0,d)$. To mitigate local trajectory collapse, where trajectories generated at different denoising steps converge to nearly identical solutions, we introduce the diversity planning losses as follows:
\begin{align}
\mathcal L_{plan} &= \mathop {\min }\limits_k d({{\tilde \tau }_{(k)}},{\tau _{{\rm{gt}}}}) - \sum\limits_{i \ne j} {||} {{\tilde \tau }_{(i)}} - {{\tilde \tau }_{(j)}}|{|_2} \nonumber \\
 &+ \sum\limits_{k = {\rm{1}}}^{K-1} {\max (0,{d_{k + {\rm{1}}}} - {d_k})} .
\end{align}
The first match loss ensures that at least one trajectory matches the ground-truth trajectory, the second diversity loss prevents trajectory collapse by encouraging different candidate trajectories across denoising steps, and the third consistency loss preserves the coarse-to-fine refinement behavior of DFM by enforcing monotonically decreasing trajectory errors throughout the denoising process.

\subsection{Perception-Oriented Trajectory Refinement}\label{POTR}
While online Group Relative Policy Optimization (GPRO) reinforcement learning proposed in~\cite{xu2026wam} and the VLA-World GPRO introduced in~\cite{wang2026learning} have designed a composite rule-based/simulator-derived reward functions to explicitly enforce safety motion planning in closed-loop control, these methods suffer from poor interpretability and their safety constraints are implicit, meanwhile, they rely heavily on known training data, resulting in poor generalization to long-tail scenarios. In this regard, inspired by CHOMP~\cite{zucker2013chomp}, we introduce an perception-oriented optimization that directly minimizes an interpretable objective function by constructing a driving risk cost map, using real-time 3D detection and BEV segmentation. It optimizes the generated multi-trajectories without additional policy training, leading to better generalization to long-tail driving scenarios.

Considering one of these predicted diverse tarjectories $\tilde \tau  = {[{x_{\rm{1}}},{y_{\rm{1}}},{x_{\rm{2}}},{y_{\rm{2}}}, \ldots ,{x_{\rm{8}}},{y_{\rm{8}}}]^T}$ as a 16-dimensional variable to be optimized, we first formulate planning optimization objective function and losses as follows: 

\begin{align}~\label{objective}
\Gamma (\tilde \tau ) = {\Gamma _{prior}} + {\Gamma _{risk}} + {\Gamma _{smooth}},
\end{align}
where the first prior cost item uses the trajectory distribution generated by Discrete Flow Matching as an optimization prior to guide the optimized trajectories to remain near the high probability driving manifold, which is expressed as: 

\begin{align}\label{prior}
{\Gamma _{prior}} = {\sum\limits_i {({\tau _i} - {{\tilde \tau }_i})} ^T}\sum\nolimits_i^{ - {\rm{1}}} {({\tau _i} - {{\tilde \tau }_i})}.
\end{align} 
And the second item is driving risk cost, that consists of two parts: semantic cost and detection cost. We utilized the results of current BEV semantic segmentation and 3D detection to construct driving risk cost map. Specifically, given the BEV projection ${B_j} = \{ {c_x},{c_y},l,w,yaw\} $ for each 3D bounding box, we compute the distance field ${D_j}({x_i},{y_i})$ for each trajectory point, and then obtain the detection-aware obstacle cost map:
\begin{align}\label{det_map}
\sum\limits_i {\sum\limits_j {\phi ({D_j}({x_i},{y_i}))} }  = \sum\limits_i {\sum\limits_j {\exp ( - \frac{{{D_j}({x_i},{y_i})}}{\sigma })} } ,
\end{align}
where $\sigma $ is distance decay parameter, which controls the rate at which the vehicle's impact on the surrounding space. The closer other objects are to the vehicle, the greater the cost and the opposite is also true. Besides, we can compute the BEV segmentation probability cost map based on BEV segmentation $\mathcal S \in {\mathbb R^{h \times w \times c}}$, assigning different costs to road information, such as drivable, lane, vehicle and sidewalk:
\begin{align}\label{seg_map}
R({x_i},{y_i}) = \sum\limits_c {\mathcal P(c|{x_i},{y_i})} .
\end{align}
The final driving risk cost map can be expressed as:
\begin{align}
{\Gamma _{risk}} = \sum\limits_c {\mathcal P(c|{x_i},{y_i})} {\omega _c} + \sum\limits_j {{\sigma _j}\phi ({D_j}({x_i},{y_i}))} ,
\end{align}
where $c$ denotes all semantic category and ${P(c|{x_i},{y_i})}$ represents the probability that the current trajectory point belongs to class $c$, and ${\omega _c}$ and ${{\sigma _j}}$ represent risk weights. As for the third item, it ensures that the trajectory remains smooth and maintains its curvature, preventing jerky movements or sharp turns, which is depicted as:
\begin{align}\label{smooth}
{\Gamma _{smooth}} = \sum\limits_i {||{\kappa _i}} |{|^2} + \sum\limits_i {||} {\tau _{i + {\rm{1}}}} - 2{\tau _i} + {\tau _{i - {\rm{1}}}}|{|^2},
\end{align}
where ${{\kappa _i}}$ is the curvature of the trajectory's tangent vector and the second part is discrete second-order difference of trajectory points. Next, we solve for the optimal path using the Levenberg-Marquardt algorithm along with Eq.~(\ref{objective}):
\begin{align}
{{\tilde \tau }^ * } = \mathop {\arg \min }\limits_\tau  \Gamma (\tilde \tau ).
\end{align}

\begin{algorithm}[t]
\caption{Multi-Trajectory Planning \& Refinement}
\label{alg:plan}
\small
\begin{algorithmic}[1]
\Require
 2D Camera visual token $\hat V$, 3D LiDAR BEV token $\hat B$, Tokenize language prompts $T$, ego states token $E$ and 
number of denoising steps $K$

\Ensure
Optimal trajectory $\tilde{\tau}_{best}^{*}$

\State Integrate multimodal discrete tokens: $Z={[V, B, T, E]}$

\State Initialize noisy trajectory tokens $x_0$

\For{$k=1,\ldots,K$}

    \State Compute posterior distribution: $p_{1|k}^{\theta}(x_{gt}\mid x_k,Z)$

    \State Sample per-trajectory tokens $x_k$

    \State Detokenize token $x_k$ into trajectory $\tilde{\tau}_{(k)}
        =
        \{(x_i,y_i)\}_{i=1}^{T_f}$

    \State Retain each trajectory from all K steps: 
    $\mathcal T
        \leftarrow
        \mathcal T
        \cup
        \{\tilde{\tau}_{(k)}\}$

\EndFor

\For{each trajectory $\tilde{\tau}_{(k)} \in \mathcal T$}

    \State Construct detection cost map
    using Eq.~(\ref{det_map})

    \State Construct semantic risk map
    using Eq.~(\ref{seg_map})

    \State Compute objective $\Gamma(\tilde{\tau}_{(k)})$ with Eq.~(\ref{prior}) and Eq.~(\ref{smooth})
  
    \State Optimize trajectory with LM: ${{\tilde \tau }_{(k)}}^* = \mathop {\arg \min }\limits_\tau  \Gamma (\tilde \tau )$
    
    \State Recalculate objective and get score:
    $score_k$
    
\EndFor

\State Select optimal trajectory: ${{\tilde \tau }_{best}}^ *  = \mathop {\arg \max }\limits_k scor{e_k}$

\State \textbf{return} $\tilde{\tau}_{best}^{*}$

\end{algorithmic}
\end{algorithm}

Similarly, we optimize each of the generated trajectories individually and obtain every optimal solution for all K trajectories $\{ \tilde \tau _{({\rm{1}})}^*,\tilde \tau _{(2)}^*, \ldots ,\tilde \tau _{(K)}^*\} $. We then recalculate the objective function $\Gamma (\tilde \tau _{(K)}^*)$ for all optimal trajectories, and define the score for each trajectory as follows:
\begin{align}
scor{e_k} = \exp ( - \Gamma (\tilde \tau _{(k)}^*)).
\end{align}
\begin{align}
{{\tilde \tau }_{best}}^ *  = \mathop {\arg \max }\limits_k scor{e_k}.
\end{align}
Thus, the lower the cost, the higher the score. We select the best trajectory based on scores.
The comprehensive and detailed inference process of multi-modal multi-tarjectory planing and refinement is provided in Fig~\ref{FIG_plan} and Alg.~\ref{alg:plan}.
% \newpage
\section{Experiments}

\subsection{Experimental Setup}

\textbf{Datasets.}~To comprehensively evaluate the performance of our method for tarjectory planning and scene perception, we conducted validation on four public datasets:
\begin{enumerate}[label=\arabic*)] 
\item \textbf{\emph{NAVSIM}}~\cite{dauner2024navsim}~is a real-world planning benchmark built upon Open-Scene, focusing on complex intention-changing scenarios while excluding trivial stationary ones. The dataset provides 360° perception from eight cameras and five LiDARs with 2 Hz annotations that include HD maps and bounding boxes.

\item \textbf{\emph{Bench2Drive}}~\cite{jia2024bench2drive}~is a closed-loop planning benchmark including several interactive scenarios (such as merging, overtaking,
yielding, emergency negotiation) in CARLA simulator. The dataset is collected from the expert model Think2Drive, comprising 1,000 clips (950 for training and 50 for testing).

\item \textbf{\emph{nuScenes}}~\cite{caesar2020nuscenes}~contains 1,000 scenes of roughly 20s duration each, and the key samples are annotated at 2Hz. Each sample consists of RGB images from 6 cameras and has ${360^ \circ }$ horizontal FOV and a 32-beam LiDAR scan. For the detection task, there are 1.4M annotated 3D bounding boxes from 10 categories. 

\item \textbf{\emph{Argoverse 2 Sensor}}~\cite{wilson2023argoverse}~contains seven surround-view cameras and a 32-beam LiDAR system along with its large detection range (200m × 200m). It contains a total of 1000 sequences, of which 700 sequences are used for training and 150 sequences for validation. We use mAP as the evaluation metric of 3D detection. Meanwhile, we selected the same six types of semantic segmentation as in CMGFA: Drivable, Bicycle, Vehicle, Stop Sign, Pedestrian, and Bollard for BEV map segmentation.

% \item \emph{Waymo Open Dataset (WOD)} is a large-scale autonomous driving dataset with both camera and LiDAR data, where WOD-E2E~\cite{xu2026wod} comprises 4,021 long-tail driving segments of 20 s each, split into 2,037/479/1,505 for train/val/test. Each segment includes eight-camera surround-views ($1920\times 1280$) video at 10 Hz along with routing inputs and ego state. Moveover, Waymo Open Dataset for perception also provides comprehensive annotations for 3D object detection and segmentation.
\end{enumerate}

\textbf{Evaluation Metrics.} For evaluation in NAVSIM dataset, we adopt closed-loop-derived metrics to assess open-loop safety and fidelity: the PDM Score (PDMS) and its extended version
(EPDMS). The PDMS integrates five criteria (Non-Collision (NC), Driving Area
Compliance (DAC),
Time-to-Collision (TTC), Comfort and Expert Progress (EP)), and EPDMS additionally includes Driving
Direction Compliance (DDC), Lane Keeping (LK) and Extended Comfort (EC).
For evaluation in Bench2Drive, we employ four core metrics:
driving score (DS), route completion (RC), infraction score
(IS), and success rate (SR). Additionally, to provide more assessment in specific aspects, we also adopt extra metrics: subdivision infraction score introduced in~\cite{jia2024bench2drive}. For evaluation in nuScenes, we employ waypoint displacement (L2) error,
Collision rate (fraction of future timestamps overlapping
with any dynamic agent), and Intersection rate (fraction
of timestamps intruding into non-drivable map regions) for open-loop planning. Moreover, in addition to mAP and mIoU for scene perception testing, the nuScenes also defines a nuScenes detection score (NDS) to capture all aspects of the nuScenes detection tasks. 
For evaluation in Argoverse 2 Sensor, the selected evaluation metric is the Intersection Over Union (IoU) for BEV segmentation. The average of the six semantic segmentations is utilized to compute the mean IoU (mIoU). 
%We use mAP as the evaluation metric of 3D detection.
%For evaluation in WOD, we report ADE and RFS, where ADE is the mean Euclidean distance over the prediction horizon and RFS evaluates a single predicted ego trajectory against multiple human-rated reference trajectories.

\textbf{Implementation Details.} All of the experiments were conducted on eight A6000 48GB GPUs, where we set four sequential training phases for tarjectory planning and select a batch size of 32 per GPU for scene perception pretraining. Perception pretraining was conducted for 20 epochs on nuScenes dataset and then applied auxiliary supervision to
fine-tune on NAVISIM dataset, including the cross-entropy loss for HD semantic maps and $L_1$ loss for 3D
bounding boxes. This stage aims to enrich our multi-modal interaction with information on various perceptions.
We use AdamW optimizer for all training stages with weight decay of 0.01. We apply trained numerical embeddings in WAM-Flow for tokenizer of ego status and detokenizer of waypoints and employ Janus-1.5B as the base VLA backbone for planning. Images
are resized with preserved aspect ratio, zero-padded
to 384×384, and encoded into 576 visual tokens and then aligns to language-token space; As for language tokens, we extend the Janus tokenizer by 20,001 numerically grounded tokens and we encode 196 BEV tokens using the same process of visual tokens. The number of 2D waypoint $T_f=8$ and the number of candidate trajectories is 5. 

\subsection{Main Results of Tarjectory Planning}

\textbf{Open-Loop Evaluation Results.}
To comprehensively evaluate our capability of multi-modality multi-tarjectory planning, we conduct open-loop experiments on NAVSIM navtest split under two types of metrics. Tab.~\ref{table_navsim-v1} summarizes the planning performance of our method on the NAVSIM test benchmark using the PDMS metric. Our method significantly surpasses previous methods, particularly on critical sub-metrics like DAC and Ego Vehicle Progress (EP) under the VLA-based manner. Our enhanced WAM-Flow, also based on the DFM Paradigm, further improves performance across
all metrics, achieving state-of-the-art results on the Navtest benchmark. Notably, its fine-grained safety reasoning yields an PDMS score of 92.2, corresponding to 90.3 of WAM-Flow, indicating high capability to drivable-area constraints. Our mehod also achieves the highest overall performance compared to traditional methods. This demonstrates the
scalability of our framework with multi-path optimization. 

\begin{figure*}[h]
    % \vspace{-0.9cm}
    \centering
    \includegraphics[width=0.95\textwidth]{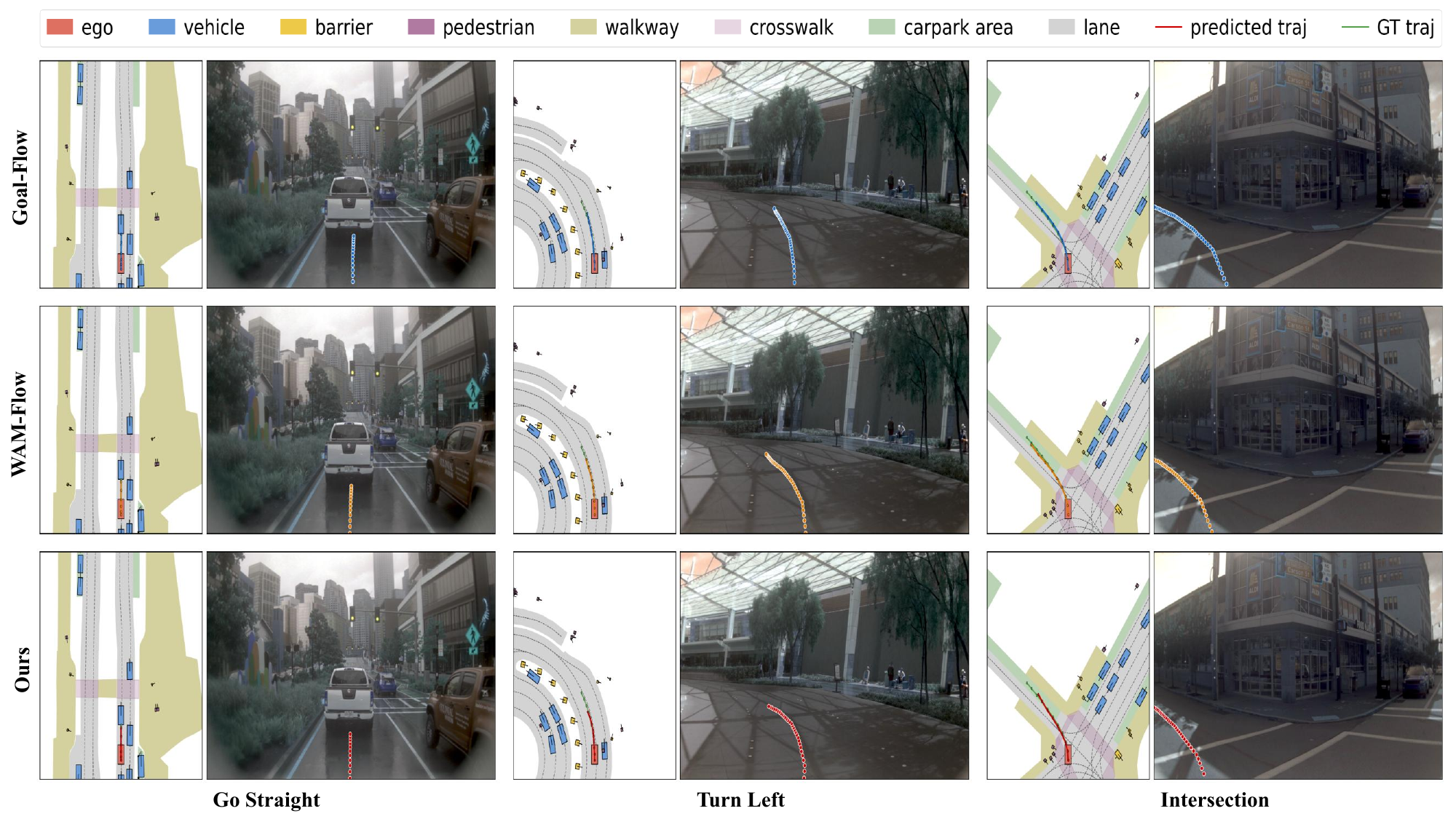} 
    % \vspace{-0.4cm}
    \caption{Visualization comparisons of our method on NAVSIM with different scenes. We conducted a comparison of our method with traditional approach: Goal-Flow and VLA-based approach: WAM-Flow. We use green to represent the ground-truth trajectory.}
    \label{FIG_navsim_show}
    % \vspace{-0.4cm}
\end{figure*} 

\begin{table*}[]

\scriptsize
\footnotesize
\setlength{\tabcolsep}{5.0pt}
\centering
\renewcommand\arraystretch{1.1}
\caption{Motion prediction performance on the NAVSIM navtest benchmark with original open-loop metrics: PDMS. Diff., Comf., and Cam denote Diffusion, Comfort and Camera input, respectively. The best and second-best VLA-based results are highlighted in bold and underline.}
\label{table_navsim-v1}
\begin{tabular}{lccccccccc}
\toprule [0.8pt]
\multicolumn{1}{l|}{Method}         & Paradigm & Backbone    & \multicolumn{1}{c|}{Input}   & NC~$\uparrow$   & DAC~$\uparrow$  & TTC~$\uparrow$  & Comf.~$\uparrow$ & \multicolumn{1}{c|}{EP~$\uparrow$}   & PDMS~$\uparrow$  \\ \midrule
\multicolumn{10}{c}{\textbf{\emph{Traditional End-to-End Manner}}}                                                                                                           \\ \midrule
\multicolumn{1}{l|}{Transfuser~\cite{chitta2022transfuser}~{\color{gray}[TPAMI2022]}}     & -        & -           & \multicolumn{1}{c|}{3×Cam + LiDAR} & 97.7 & 92.8 & 92.8 & 100   & \multicolumn{1}{c|}{79.2} & 84.0  \\
\multicolumn{1}{l|}{GoalFlow~\cite{xing2025goalflow}~{\color{gray}[CVPR2025]}}    & -        & -           & \multicolumn{1}{c|}{3×Cam + LiDAR} & 98.4 & 98.3 & 94.6 & 100   & \multicolumn{1}{c|}{85.0} & 90.3 \\
\multicolumn{1}{l|}{VADv2~\cite{jiang2024vadv2}~{\color{gray}[ICLR2024]}}          & -        & -           & \multicolumn{1}{c|}{6×Cam}     & 97.2 & 89.1 & 91.6 & 100   & \multicolumn{1}{c|}{76.0} & 80.9  \\
\multicolumn{1}{l|}{DiffusionDrive~\cite{liao2025diffusiondrive}~{\color{gray}[CVPR2025]}} & Diff.    & -           & \multicolumn{1}{c|}{3×Cam + LiDAR} & 98.2 & 96.0 & 94.8 & 100   & \multicolumn{1}{c|}{82.2} & 88.1  \\
\multicolumn{1}{l|}{Artemis~\cite{feng2025artemis}~{\color{gray}[RAL2025]}}        & Diff.    & -           & \multicolumn{1}{c|}{6×Cam}     & 98.3 & 95.1 & 94.3 & 99.8  & \multicolumn{1}{c|}{81.4} & 87.0  \\ 
\multicolumn{1}{l|}{SafeDrive~\cite{kim2026safedrive}~{\color{gray}[CVPR2025]}}        & -    & -           & \multicolumn{1}{c|}{3×Cam + LiDAR}     & 99.5 & 99.0 & 97.2 & 84.3  & \multicolumn{1}{c|}{100} & 91.6  \\
\midrule
\multicolumn{10}{c}{\textbf{\emph{VLA-Based End-to-End Manner}}}                                                                                                             \\ \midrule
\multicolumn{1}{l|}{DrivingGPT~\cite{chen2025drivinggpt}~{\color{gray}[ICCV2025]}}     & AR       & LLaMA2-7B   & \multicolumn{1}{c|}{1×Cam}     & 98.1 & 90.7 & 94.9 & 95.6  & \multicolumn{1}{c|}{79.7} & 82.4  \\
\multicolumn{1}{l|}{FSDrive~\cite{zeng2026futuresightdrive}~{\color{gray}[NeurIPS2026]}}        & AR       & Qwen2-VL-2B & \multicolumn{1}{c|}{6×Cam}     & 98.2 & 93.8 & 93.3 & \underline{99.9}  & \multicolumn{1}{c|}{80.1} & 85.1  \\
\multicolumn{1}{l|}{AutoVLA~\cite{zhou2026autovla}~{\color{gray}[NeurIPS2026]}}        & AR       & Qwen2.5-3B  & \multicolumn{1}{c|}{3×Cam}     & 98.4 & 95.6 & \underline{98.0} & \underline{99.9}  & \multicolumn{1}{c|}{81.9} & 89.1  \\
\multicolumn{1}{l|}{Epona~\cite{zhang2025epona}~{\color{gray}[ICCV2025]}}          & AR+Diff. & DiT-2.5B    & \multicolumn{1}{c|}{1×Cam}     & 97.9 & 95.1 & 93.8 & \underline{99.9}  & \multicolumn{1}{c|}{80.4} & 86.2  \\
\multicolumn{1}{l|}{WAM-Flow~\cite{xu2026wam}~{\color{gray}[CVPR2026]}}       & DFM      & Janus-1.5B  & \multicolumn{1}{c|}{1×Cam}     & \textbf{99.2} & \underline{98.3} & 97.0 & 99.7  & \multicolumn{1}{c|}{\underline{82.3}} & \underline{90.3}  \\ \midrule
\rowcolor{gray!10}\multicolumn{1}{l|}{Ours}           & DFM        & Janus-1.5B           & \multicolumn{1}{c|}{Cam + LiDAR}       & \underline{99.0}  & \textbf{98.5}  & \textbf{98.3}  & \textbf{100}   & \multicolumn{1}{c|}{\textbf{88.6}}  & \textbf{92.2}   \\ \bottomrule [0.8pt]
\end{tabular}
\end{table*}

\begin{table*}[!t]
\scriptsize
\footnotesize
\setlength{\tabcolsep}{7.0pt}
\centering
\renewcommand\arraystretch{1.1}
\caption{Motion prediction performance on the NAVSIM navtest benchmark with extended open-loop metrics: EPDMS. "Camera" and "LiDAR" denote Camera and LiDAR inputs, respectively. The best and second-best results are highlighted in bold and underline.}
\label{table_navsim-v2}
\begin{tabular}{l|c|ccccccc|c}
\toprule [0.8pt]
Method         & Input & NC~$\uparrow$   & DAC~$\uparrow$  & EP~$\uparrow$   & TTC~$\uparrow$  & DDC~$\uparrow$  & LK~$\uparrow$   & EC~$\uparrow$   & EPDMS~$\uparrow$ \\ \midrule
VADv2~\cite{jiang2024vadv2}~{\color{gray}[ICLR2024]}          & Camera     & 97.3 & 91.7 & 77.6 & 92.7 & 98.2 & 66.0 & 97.4 & 76.6  \\
TransFuser~\cite{chitta2022transfuser}~{\color{gray}[TPAMI2022]}     & Camera \& LiDAR  & 97.7 & 92.8 & 79.2 & 92.8 & 98.3 & 67.6 & 95.3 & 77.8  \\
DiffusionDrive~\cite{liao2025diffusiondrive}~{\color{gray}[CVPR2025]} & Camera \& LiDAR  & 98.2 & 96.2 & \underline{87.6} & 97.3 & \underline{98.6} & 97.0 & \underline{98.4} & 84.0  \\
GaussianFusion~\cite{liu2026gaussianfusion}~{\color{gray}[NeurIPS2026]} & Camera \& LiDAR  & \underline{98.3} & \underline{97.3} & 87.5 & \underline{97.4} & \textbf{99.0} & \underline{97.4} & 98.3 & \underline{85.0}  \\ 
\midrule
\rowcolor{gray!10}Ours           & Camera \& LiDAR     & \textbf{99.0}  & \textbf{98.5}  & \textbf{88.6}  & \textbf{98.3}  & \textbf{99.0}  & \textbf{97.6}  & \textbf{98.7}  & \textbf{87.0}   \\ \bottomrule [0.8pt]
\end{tabular}
\end{table*}

\newpage
\begin{table*}[]
\scriptsize
\footnotesize
\setlength{\tabcolsep}{3.5pt}
\centering
\renewcommand\arraystretch{1.1}
\caption{Motion prediction performance on Bench2Drive benchmark with core metrics and subdivision infraction scores. DS, RC, IS, SR correspond to the Driving Score, Route Completion, Infraction Score, and Success Rate. CP, CV, CL, RL, SS, OR, AB, YEV correspond to the Collision with a Pedestrian,
Collision with another Vehicle, Collision with Layout, Red Light infractions, Stop Sign infractions, Off-Road infractions, Agent Blocked,
and failure to Yield to Emergency Vehicles infractions. All comparison results are from VLR-Driver~\cite{kong2025vlr}. The best and second-best VLM/VLA-based results are highlighted in bold and underline.}
\label{table_benchdrive}
\begin{tabular}{lccccccccccccc}
\toprule [0.8pt]
\multicolumn{1}{c|}{\multirow{2}{*}{Method}} & \multicolumn{1}{c|}{\multirow{2}{*}{Input Modality}} & \multicolumn{4}{c|}{Core Metrics~$\uparrow$}                  & \multicolumn{8}{c}{Subdivision Infraction Score~$\downarrow$}              \\
\multicolumn{1}{c|}{}                        & \multicolumn{1}{c|}{}                          & DS    & RC    & IS    & \multicolumn{1}{c|}{SR}    & CP    & CV    & CL    & RL    & SS    & OR    & AB    & YEV   \\ \midrule
\multicolumn{14}{c}{\textbf{\emph{Traditional End-to-End Manner}}}                                                                                                                                                                 \\ \midrule
\multicolumn{1}{l|}{TransFuser~\cite{chitta2022transfuser}~{\color{gray}[TPAMI2022]} }              & \multicolumn{1}{c|}{Camera \& LiDAR}                       & 37.18 & 68.14 & 0.51  & \multicolumn{1}{c|}{9.09}  & 0.96  & 13.24 & 8.71  & 0.00  & 0.96  & 0.32  & 2.58  & 0.32  \\
\multicolumn{1}{l|}{ThinkTwice~\cite{jia2023think}~{\color{gray}[CVPR2023]} }              & \multicolumn{1}{c|}{Camera \& LiDAR}                       & 58.79 & 74.35 & 0.77  & \multicolumn{1}{c|}{29.54} & 0.30  & 5.76  & 0.91  & 0.00  & 0.91  & 0.05  & 0.91  & 0.30  \\

\multicolumn{1}{l|}{TCP~\cite{wu2022trajectory}~{\color{gray}[NeurIPS2022]} }                  & \multicolumn{1}{c|}{Camera}                       & 56.28 & 83.57 & 0.65  & \multicolumn{1}{c|}{25.00} & 0.26  & 5.46 & 5.46  & 0.00  & 0.52  & 0.22  & 0.78  & 0.00  \\

\multicolumn{1}{l|}{NEAT~\cite{chitta2021neat}~{\color{gray}[ICCV2021]} }              & \multicolumn{1}{c|}{Camera}                       & 30.86 & 55.35 & 0.55  & \multicolumn{1}{c|}{6.81} & 1.08  & 9.87  & 5.57  & 0.20  & 1.33  & 0.41  & 2.01  & 0.27  \\ \midrule
\multicolumn{14}{c}{\textbf{\emph{VLM/VLA-Based End-to-End Manner}}}                                                                                                                                                                   \\ \midrule
\multicolumn{1}{l|}{LMDrive~\cite{shao2024lmdrive}~{\color{gray}[CVPR2024]} }                 & \multicolumn{1}{c|}{Camera \& LiDAR \& Language}                     & 24.76 & 33.02 & \textbf{0.90}  & \multicolumn{1}{c|}{13.63} & 1.14  & 2.86  & 2.29  & \textbf{0.00}  & 0.57  & \underline{0.05}  & 3.44  & 0.57  \\
\multicolumn{1}{l|}{LeapAD~\cite{mei2024continuously}~{\color{gray}[NeurIPS2024]} }                  & \multicolumn{1}{c|}{Camera \& Language}                       & 55.18 & 77.45 & 0.71  & \multicolumn{1}{c|}{36.36} & \underline{0.69}  & 5.07  & 1.15  & 0.20  & 0.91  & 0.08  & 1.47  & \underline{0.27}  \\
\multicolumn{1}{l|}{VLR-Driver~\cite{kong2025vlr}~{\color{gray}[ICCV2025]} }              & \multicolumn{1}{c|}{Camera \& Language}                       & \underline{75.01} & \underline{86.09} & \underline{0.87}  & \multicolumn{1}{c|}{\underline{50.00}} & 0.72  & \underline{2.83}  & \textbf{0.48}  & \textbf{0.00}  & \underline{0.48}  & \textbf{0.03}  & \textbf{0.25}  & \textbf{0.25}  \\ \midrule
\rowcolor{gray!10}\multicolumn{1}{l|}{Ours}                    & \multicolumn{1}{c|}{Camera \& LiDAR \& Language}                     & \textbf{77.45} & \textbf{88.23} & \textbf{0.90} & \multicolumn{1}{c|}{\textbf{56.00}} & \textbf{0.55} & \textbf{2.82} & \underline{0.50} & \textbf{0.00} & \textbf{0.47} & \textbf{0.03} & \underline{0.44} & 0.32 \\ \bottomrule [0.8pt]
\end{tabular}
\end{table*}

\begin{table*}[t]
\scriptsize
\footnotesize
\setlength{\tabcolsep}{6.2pt}
\centering
\renewcommand\arraystretch{1.1}
\caption{Motion prediction performance on nuScenes benchmark with open-loop metrics. Results are highlighted in bold and underline for the best and second-best performance among VLM/VLA-based methods. All comparison results follow OmniDrive and are from SpaceDrive~\cite{li2026spacedrive}.}
\label{table_nuscenes}
\begin{tabular}{lcccccccccccccc}
\toprule [0.8pt]
\multicolumn{1}{c|}{\multirow{2}{*}{Method}} & \multicolumn{2}{c|}{Ego Status}    & \multicolumn{4}{c|}{L2 (m)~$\downarrow$}                    & \multicolumn{4}{c|}{Collision (\%)~$\downarrow$}            & \multicolumn{4}{c}{Intersection (\%)~$\downarrow$} \\
\multicolumn{1}{c|}{}                        & BEV & \multicolumn{1}{c|}{Planner} & 1s   & 2s   & 3s   & \multicolumn{1}{c|}{Avg.} & 1s   & 2s   & 3s   & \multicolumn{1}{c|}{Avg.} & 1s      & 2s      & 3s      & Avg.    \\ \midrule
\multicolumn{15}{c}{\textbf{\emph{Traditional Modular End-to-End Paradigm}}}                                                                                                                                                                \\ \midrule
\multicolumn{1}{l|}{UniAD~\cite{hu2023planning}~{\color{gray}[CVPR2023]}}                   & \ding{51}   & \multicolumn{1}{c|}{\ding{51}}       & 0.20 & 0.42 & 0.75 & \multicolumn{1}{c|}{0.46} & 0.02 & 0.25 & 0.84 & \multicolumn{1}{c|}{0.37} & 0.20    & 1.33    & 3.24    & 1.59    \\
\multicolumn{1}{l|}{VAD-Base~\cite{jiang2023vad}~{\color{gray}[ICCV2023]}}                & \ding{51}   & \multicolumn{1}{c|}{\ding{51}}       & 0.17 & 0.34 & 0.60 & \multicolumn{1}{c|}{0.37} & 0.04 & 0.27 & 0.67 & \multicolumn{1}{c|}{0.33} & 0.21    & 2.13    & 5.06    & 2.47    \\
\multicolumn{1}{l|}{BEV-Planner~\cite{li2024ego}~{\color{gray}[CVPR2024]}}             & \ding{51}   & \multicolumn{1}{c|}{\ding{51}}       & 0.16 & 0.32 & 0.57 & \multicolumn{1}{c|}{0.35} & 0.00 & 0.29 & 0.73 & \multicolumn{1}{c|}{0.34} & 0.35    & 2.62    & 6.51    & 3.16    \\
\multicolumn{1}{l|}{UAD~\cite{guo2025end}~{\color{gray}[TPAMI2025]}}                     & \ding{51}   & \multicolumn{1}{c|}{\ding{51}}       & 0.13 & 0.28 & 0.48 & \multicolumn{1}{c|}{0.30} & 0.00 & 0.19 & 0.16 & \multicolumn{1}{c|}{0.27} & 0.13    & 1.08    & 2.89    & 1.37    \\ \midrule
\multicolumn{15}{c}{\textbf{\emph{VLM/VLA-Based End-to-End Paradigm}}}                                                                                                                                                                          \\ \midrule
\multicolumn{1}{l|}{RDA-Driver~\cite{huang2024making}~{\color{gray}[ECCV2024]}}              & \ding{51}   & \multicolumn{1}{c|}{\ding{51}}       & 0.23 & 0.73 & 1.54 & \multicolumn{1}{c|}{0.80} & \textbf{0.00} & \textbf{0.13} & 0.83 & \multicolumn{1}{c|}{\underline{0.32}} & -       & -       & -       & -       \\
\multicolumn{1}{l|}{DriveVLM~\cite{tian2024drivevlm}~{\color{gray}[CoRL2024]}}                & -   & \multicolumn{1}{c|}{\ding{51}}       & 0.18 & 0.34 & 0.68 & \multicolumn{1}{c|}{0.40} & 0.10 & 0.22 & \textbf{0.45} & \multicolumn{1}{c|}{0.27} & -       & -       & -       & -       \\
\multicolumn{1}{l|}{ORION~\cite{fu2025orion}~{\color{gray}[ICCV2025]}}                   & \ding{51}   & \multicolumn{1}{c|}{-}       & 0.17 & 0.31 & 0.55 & \multicolumn{1}{c|}{0.34} & 0.05 & 0.25 & 0.80 & \multicolumn{1}{c|}{0.37} & -       & -       & -       & -       \\
\multicolumn{1}{l|}{OmniDrive~\cite{wang2025omnidrive}~{\color{gray}[CVPR2025]}}               & -   & \multicolumn{1}{c|}{-}       & \textbf{0.14} & \underline{0.29} & 0.55 & \multicolumn{1}{c|}{0.33} & \textbf{0.00} & \textbf{0.13} & 0.78 & \multicolumn{1}{c|}{0.30} & 0.56    & 2.48    & 5.96    & 3.00    \\
\multicolumn{1}{l|}{SpaceDrive~\cite{li2026spacedrive}~{\color{gray}[CVPR2026]}}              & -   & \multicolumn{1}{c|}{-}       & 0.15 & \underline{0.29} & \underline{0.51} & \multicolumn{1}{c|}{0.32} & 0.04 & 0.18 & \underline{0.49} & \multicolumn{1}{c|}{\underline{0.23}} & \textbf{0.22}    & \textbf{0.80}    & \underline{2.79}    & \textbf{1.27}    \\ \midrule
\rowcolor{gray!10}\multicolumn{1}{l|}{Ours}                    & \ding{51}   & \multicolumn{1}{c|}{\ding{51}}       & \textbf{0.14} & \textbf{0.27} & \textbf{0.50} & \multicolumn{1}{c|}{\textbf{0.30}} & \underline{0.03} & \underline{0.15} & \textbf{0.45} & \multicolumn{1}{c|}{\textbf{0.21}} & \underline{0.31}    & \underline{1.00}    & \textbf{2.54}    & \textbf{1.27}    \\ \bottomrule [0.8pt]
\end{tabular}
\vspace{-0.4cm}
\end{table*}

Besides, we evaluate performance under the expanded EPDMS metric, and the results are presented in Tab.~\ref{table_navsim-v2}. Note that EPDMS poses a stricter challenge than PDMS by incorporating more slight driving criteria. Our method achieves 87.0 EPDMS score,
demonstrating its significant superior performance over most of previous conventional methods.
Compared to TransFuser, our method surpasses it by 9.2 EPDMS while employing a multi-modal fusion strategy. We also outperforms GaussianFusion, which follows a dual-branch fusion pipeline tailored to the planning-centric task, with a 2.0 EPDMS improvement. As shown in Fig. \ref{FIG_navsim_show}, we present a qualitative comparison among GoalFlow, WAM-FLow, and our proposed approach in representative turning scenarios. Our method generates a diverse distribution of feasible trajectories, effectively modeling multiple plausible driving intentions and future behaviors. The results demonstrate its superiority over conventional VLM-based planning approaches, which often converge to a limited set of conservative solutions due to mode collapse, leading to insufficient exploration of the trajectory space.

\textbf{Closed-Loop Evaluation Results.} To establish a more comprehensive and reliable assessment of tarjectory planning performance, we further present close-loop results on the Bench2Drive dataset containing 44 diverse interactive scenarios, as shown in Tab.~\ref{table_benchdrive}. Our method achieves the SOTA performance across all core metrics, consistently surpassing existing VLM/VLA-based methods. Compared to VLR-Driver, it improves Driving Score by 2.44\%, Route Completion by 2.14\%, Infraction Score by 0.03\%, and Success Rate by 6.00\%. 
These results indicate that our method consistently generates high-quality trajectories, achieving better task success while maintaining superior comfort and stability in complex environments. Moreover, in Subdivision Infraction evaluation, we achieves the best performance across most metrics. In particular, it reduces the incidence of pedestrian collisions and vehicle collisions by 0.14 and 0.01, respectively, indicating enhanced scene understanding and more reliable decision-making in safety-critical situations.
This demonstratesthat our model achieves consistent and robust driving performance across flexible scenarios.

\textbf{Long-Horizon Planning Results.}
To evaluate long-horizon tarjectory planning ability of our method, we then conducted the open-loop planning experiment on the nuScenes dataset, and the results are shown in Tab.~\ref{table_nuscenes}. 
We achieve the highest diversity average metric across all prediction timestamps, respectively, clearly outperforming all baselines. The lowest L2 error (0.30\%) verifies the effectiveness of our multi-modal trajectory planning and optimization mechanism, which refines candidate trajectories toward more accurate and realistic driving behaviors, thereby achieving better alignment with expert demonstrations. Additionally, we report the lowest average collision rate (0.21\%) and intersection rate (1.27\%), significantly better than
SpaceDrive (0.23\%~\&~1.27\%) and OmniDrive (0.30\%~\&~3.00\%). These results demonstrate that our multi-modal interaction-aware planning framework effectively generates and optimizes diverse trajectory hypotheses, enabling more accurate driving predictions, and yielding better long-term planning performance.

%We also present performance comparisons on the WOD-E2E test set, which specifically handles challenging long-tail driving scenarios, the results are shown in Tab.~\ref{table_wod-e2e}. Our method obtains the highest RFS (8.000) and ADE@3s (8.000) metrics. Note that the AutoVLA achieve substantially lower for ADE@5s metric, likely due to its reinforcement fine-tuning. The superior RFS and lowest ADE@3s further verify that the proposed multi-modal interaction and trajectory optimization paradigm effectively bridges high-level semantic reasoning and low-level motion planning, yielding more human-aligned behaviors and more accurate long-term trajectory predictions in real-time and safety-critical driving scenarios.

\begin{table}[]
\scriptsize
\footnotesize
\setlength{\tabcolsep}{4.2pt}
\centering
\renewcommand\arraystretch{1.1}
\caption{Scene perception performance comparison of 3D object detection on nuScenes test and val set. "L" and "C" indicate LiDAR and camera.}
\label{table_perception-nuscene}
\begin{tabular}{l|c|cc|cc}
\toprule [0.8pt]
\multicolumn{1}{c|}{\multirow{2}{*}{Method}} & \multirow{2}{*}{Modality} & \multicolumn{2}{c|}{validation} & \multicolumn{2}{c}{test} \\
\multicolumn{1}{c|}{}                        &                           & mAP~$\uparrow$          & NDS~$\uparrow$         & mAP~$\uparrow$       & NDS~$\uparrow$      \\ \midrule
CenterPoint~\cite{yin2021center}                                  & L                         & 59.6         & 66.8             & 60.3      & 67.3         \\
GeoFormer~\cite{jin2025geoformer}                                     & L+C                         & 68.2         & 72.4             & 69.8      & 73.7         \\ 
TransFusion~\cite{bai2022transfusion}                                  & L+C                       & 67.5         & 71.3             & 68.9      & 71.6         \\
BEVFusion~\cite{liu2023bevfusion}                                    & L+C                       & 67.9         & 71.0             & 69.2      & 71.8         \\
BEVFormer~\cite{li2024bevformer}                                    & L+C                       & \underline{71.2}         & \underline{73.2}             & \underline{72.9}      & \underline{74.1}         \\ \midrule
\rowcolor{gray!10}Ours                                         & L+C                       & \textbf{73.7}         & \textbf{77.2}             & \textbf{74.0}      & \textbf{78.0}         \\ \bottomrule [0.8pt]
\end{tabular}
\end{table}

\begin{table}[]
\scriptsize
\footnotesize
\setlength{\tabcolsep}{2.2pt}
\centering
\renewcommand\arraystretch{1.1}
\caption{Scene perception performance comparison of 3D object detection on Argoverse 2. All comparison results are all from GeoFormer~\cite{jin2025geoformer}.}
\label{table_arg_detection}
\begin{tabular}{l|c|ccccc}
\toprule [0.8pt]
\multicolumn{1}{c|}{Method} & mAP           & Veh.          & Bus           & Ped.          & Truck         & Bicyclist     \\ \midrule
CenterPoint~\cite{yin2021center}~{\color{gray}[CVPR2021]}                & 22.0          & 67.6          & 38.9          & 46.5          & 22.1          & 20.1          \\
VoxelNeXt~\cite{chen2023voxelnext}~{\color{gray}[CVPR2023]}                  & 30.7          & 72.7          & 38.8          & 63.2          & 16.9          & 32.4          \\
HEDNet~\cite{zhang2023hednet}~{\color{gray}[NeurIPS2023]}                     & 37.1          & 78.2          & 47.7          & 67.6          & 21.6          & 38.7          \\
FSDv2~\cite{fan2024fsd}~{\color{gray}[TPAMI2024]}                      & 37.6          & 77.0          & 47.6          & 70.5          & 24.0          & 45.9          \\
SAFDNet~\cite{zhang2024safdnet}~{\color{gray}[CVPR2024]}                    & 39.7          & \underline{78.5}    & 49.4          & 70.7          & 23.6          & 42.7          \\
GeoFormer~\cite{jin2025geoformer}~{\color{gray}[ICCV2025]}                 & \underline{41.7}    & 77.4          & \textbf{50.7} & \underline{73.7}    & \underline{24.6}    & \textbf{48.0} \\ \midrule
\rowcolor{gray!10}Ours                       & \textbf{43.5} & \textbf{79.0} & \underline{49.6}    & \textbf{75.9} & \textbf{26.3} & \underline{45.1} \\ \bottomrule [0.8pt]  
\end{tabular}
\vspace{-0.3cm}
\end{table}

% \begin{table}[]
% \scriptsize
% \footnotesize
% \setlength{\tabcolsep}{4.2pt}
% \centering
% \renewcommand\arraystretch{1.1}
% \caption{Motion prediction performance comparison on WOD-E2E test set. }
% \label{table_wod-e2e}
% \begin{tabular}{l|c|ccc}
% \toprule [0.8pt]
% \multicolumn{1}{c|}{Method} & Backbone     & RFS~$\uparrow$   & ADE (5s)~$\downarrow$ & ADE (3s)~$\downarrow$ \\ \midrule
% Open-LLaMA                  & LLaMA-Vision & 7.429 & 3.217    & 1.314    \\
% NaiveEMMA                   & Gemini       & 7.528 & 3.018    & 1.320    \\
% AutoVLA                     & Qwen2.5-VL   & 7.557 & 2.558    & 1.351    \\
% dVLM-AD                     & LLaDA-V      & 7.633 & 3.022    & 1.285    \\ \midrule
% Ours                        & -            & 8.000 & 8.000    & 8.000 \\ \bottomrule [0.8pt]
% \end{tabular}
% \end{table}

\begin{figure*}[t]
    % \vspace{-0.9cm}
    \centering
    \includegraphics[width=\textwidth]{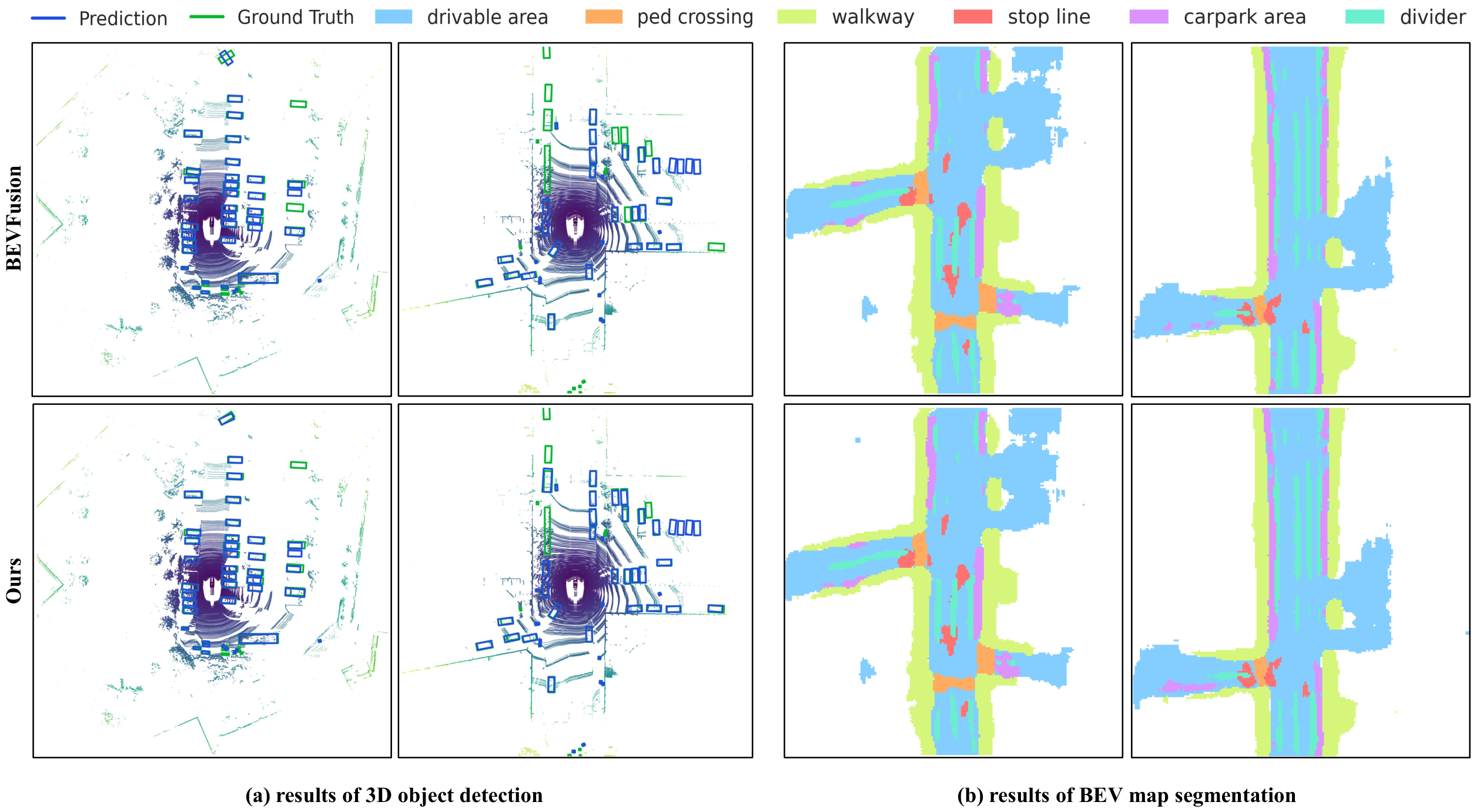} 
    % \vspace{-0.4cm}
    \caption{Visualization comparisons of both 3D object detection and map segmentation tasks on NuScenes dataset. We conducted a comparison of our method with BEVFusion. In object detection task, we use blue and green to represent the prediction boxes and the ground-truth boxes.}
    \label{FIG_nuscenes_show}
    \vspace{-0.2cm}
\end{figure*}

\subsection{Main Results of Scene Perception}

\textbf{3D Object Detection.}
In this section, we evaluate the 3D detection performance of our multi-modal interaction framework and compare it with previous state-of-the-art methods. As shown in Tab.~\ref{table_perception-nuscene}, our method achieves the best performance on both the validation and test sets. Under the multi-modal setting, our method obtains 73.7\% mAP and 77.2\% NDS on the validation set, and 74.0\% mAP and 78.0\% NDS on the test set, outperforming the previous best method, BEVFormer, by 2.5 and 4.0 in mAP and NDS on the validation set, respectively. On the test set, our method further improves the mAP and NDS by 1.1 and 3.9, respectively. These results demonstrate the effectiveness of our multi-modal interaction in exploiting fusion information for scene perception.
The detection results in Fig.~\ref{FIG_nuscenes_show}(a) can also observe that we produce more stable real-time detection, which aids in the final tarjectory planning.
We argue that effective multi-modal interaction constitutes a key component of AD systems.

As shown in Tab.~\ref{table_arg_detection}, our multi-modal interaction method also achieves the best overall performance on Argoverse 2 dataset, reaching 43.5 mAP, surpassing GeoFormer by 1.8 and SAFDNet by 3.8. Notably, our method consistently improves the detection of major object categories, achieving 79.0, 75.9, and 26.3 average precision for vehicles, pedestrians, and trucks, respectively, which are the best results among all compared methods. The substantial gains on pedestrians and trucks indicate that our multi-modal interaction effectively leverages complementary information across different modalities. Although GeoFormer performs slightly better on bicyclists, our method still maintains competitive performance with 45.1 AP. Overall, these results demonstrate that the proposed multi-modal interaction strategy provides more effective cross-modal feature representations and substantially enhances 3D detection.

\begin{table}[]
\scriptsize
\footnotesize
\setlength{\tabcolsep}{3pt}
\centering
\renewcommand\arraystretch{1.1}
\caption{Scene perception comparison of BEV segmentation on the nuScenes val set using the mIoU metric. "L" and "C" indicate LiDAR and camera.}
\label{table_segment-nuscene}
\begin{tabular}{c|cccc|>{\columncolor{gray!10}}c}
\toprule[0.8pt]
Method     & \begin{tabular}[c]{@{}c@{}}PointPillars\\ \cite{vora2020pointpainting}\end{tabular} & \begin{tabular}[c]{@{}c@{}}CenterPoint\\ \cite{yin2021center}\end{tabular} & \begin{tabular}[c]{@{}c@{}}MVP\\ \cite{yin2021multimodal}\end{tabular} & \begin{tabular}[c]{@{}c@{}}BEVFusion\\ \cite{liu2023bevfusion}\end{tabular} & Ours \\
\midrule
Modality    & L & L & C+L & C+L & C+L \\ \midrule
Drivable    & 72.0 & 75.6 & 76.1 & \underline{85.5} & \textbf{88.2} \\
Ped.Cross.  & 43.1 & 48.4 & 48.7 & \underline{60.5} & \textbf{73.4} \\
Walkway     & 53.1 & 57.5 & 57.0 & \underline{67.6} & \textbf{82.1} \\
Stop Line   & 29.7 & 36.5 & 36.9 & \underline{52.0} & \textbf{60.9} \\
Carpark     & 27.7 & 31.7 & 33.0 & \underline{57.0} & \textbf{62.4} \\
Divider     & 37.5 & 41.9 & 42.2 & \underline{53.7} & \textbf{60.0} \\
\midrule
Average        & 43.8 & 48.6 & 49.0 & \underline{62.7} & \textbf{71.1} \\
\bottomrule[0.8pt]
\end{tabular}
\end{table}

\begin{table}[]
\scriptsize
\footnotesize
\setlength{\tabcolsep}{3.5pt}
\centering
\renewcommand\arraystretch{1.1}
\caption{Scene perception comparison of BEV segmentation on Argoverse 2 dataset. All comparison results are all from CMGFA~\cite{kuang2024cmgfa}.}
\label{table_segment-arg}
\begin{tabular}{c|ccc|>{\columncolor{gray!10}}c}
\toprule[0.8pt]
Method     & Simple-BEV~\cite{harley2023simple} & BEVFusion~\cite{liu2023bevfusion} & CMGFA~\cite{kuang2024cmgfa}         & Ours          \\ \midrule
Drivable   & 72.3       & 80.8      & \underline{85.9}    & \textbf{87.5} \\
Bicycle    & 50.1       & 56.0      & \underline{60.8}    & \textbf{63.8} \\
Vehicle    & 53.2       & 63.2      & \underline{69.3}    & \textbf{70.7} \\
Stop sign  & 42.3       & 52.2      & \textbf{56.8} & \underline{55.1}    \\
Pedestrain & 45.1       & 51.2      & \underline{55.2}    & \textbf{59.4} \\
Bollard    & 42.1       & 48.3      & \textbf{52.1} & \underline{46.5}    \\ \midrule
Average    & 50.9       & 58.6      & \underline{63.3}    & \textbf{63.8} \\ \bottomrule[0.8pt]
\end{tabular}
\vspace{-0.3cm}
\end{table}

\textbf{BEV Map Segmentation.}~As described in Tab.~\ref{table_segment-nuscene}, we further compare our multi-modal interaction with state-of-the-art 3D
perception models on the semantic-centric BEV map segmentation task on nuScenes dataset, where we report the Intersection-over-Union
(IoU) on six background classes and the class-averaged mean IoU as our evaluation metric. Following the evaluation protocol, each semantic category is treated as an independent binary segmentation problem because different classes may overlap spatially (\emph{e.g.}, parking areas and drivable regions). The reported score corresponds to the highest IoU obtained across different threshold settings. As a result, we outperform LiDAR-only baselines by 22.5-27.3 mIoU, and our model also improves upon existing camera-LiDAR fusion-based methods by 8.4 mIoU. In particular, our method outperforms BEVFusion by 8.4 mIoU, achieving an average mIoU of 71.1\%. Through cross-modal interaction, the model leverages dense visual observations from camera images to enrich semantic perception while utilizing LiDAR measurements to preserve accurate spatial structures, resulting in improved scene representation quality. As depicted in Fig.~\ref{FIG_nuscenes_show}(b), our predicted BEV maps exhibit higher spatial consistency, where large semantic regions remain smooth and continuous while fine-grained structures are better preserved.
%The fused BEV representations provide more accurate environmental semantics and spatial context, allowing the following planning module to generate trajectories that are safer, more feasible, and better aligned with the surrounding scene.

Tab.~\ref{table_segment-arg} reports the BEV map segmentation results on the Argoverse 2 dataset. Our method achieves the best overall performance with an average IoU of 63.8\%, outperforming CMGFA by 0.5 and significantly surpassing Simple-BEV and BEVFusion by 12.9 and 5.2, respectively. These improvements indicate that the proposed multi-modal interaction strategy effectively integrates complementary information from different modalities, leading to more accurate BEV semantic understanding and enhanced scene representation. The proposed multi-modal interaction can capture both global scene context and fine-grained semantic structures, thereby improving BEV map perception.

\subsection{Ablation Studies \& Extra Analysis}

\subsubsection{Ablations of Multi-Trajectory Planning \& Optimization}

\textbf{Impact of Multi-Trajectory Planning loss.}~As shown in Tab.~\ref{table_abl_multi_traj}, loss function design in Eq.~(\ref{loss_plan}) has a significant impact on multi-trajectory planning performance. Starting from the matching loss $\mathcal L_{match}$, the model achieves a PDMS of 86.9, while introducing the diversity loss $\mathcal L_{diversity}$ further improves PDMS to 87.5, indicating that explicitly encouraging trajectory diversity helps the model generate more effective driving alternatives. With the additional consistency loss $\mathcal L_{cons.}$, the overall planning loss achieves the best performance, substantially improving PDMS to 92.2 and EP from 85.4 to 88.6, yielding consistent gains in DAC and TTC, leading to more reliable planning performance.

\begin{figure}[ht]
    % \vspace{-0.9cm}
    \centering
    \includegraphics[width=0.49\textwidth]{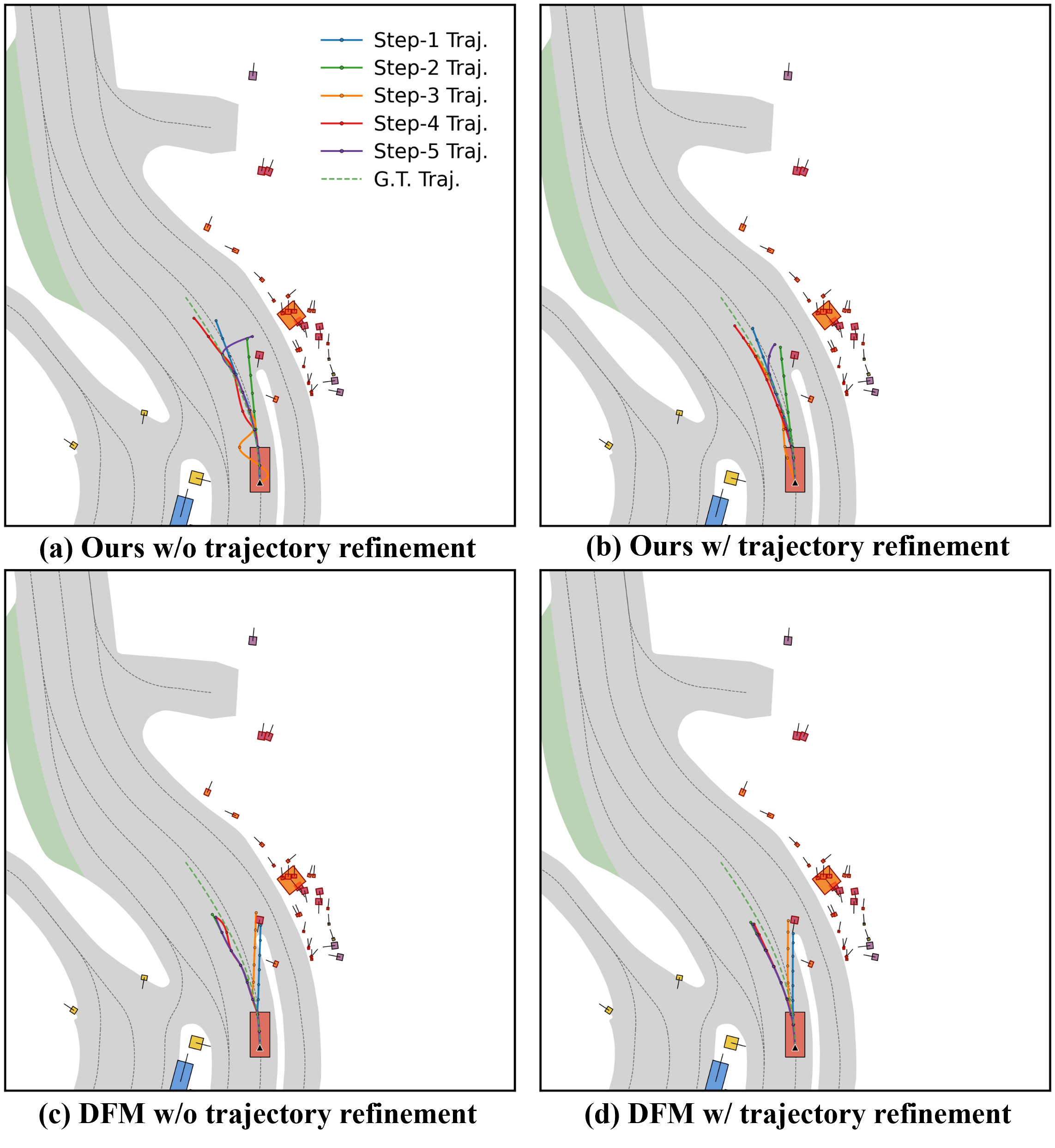} 
    \vspace{-0.4cm}
    \caption{Ablation comparisons of trajectory optimization on NAVSIM dataset. We conducted a comparison of our multi-trajectory generation and DFM-based manner along with optimization module.}
    \label{FIG_abl_opt}
    % \vspace{-0.4cm}
\end{figure}

\textbf{Adaptability of VLA Foundations.} We further evaluate the adaptability of our framework across different VLA foundations. As shown in Tab.~\ref{table_abl_vlm}, our method achieves consistently strong performance on Qwen2.5-3B, LLaVA-7B, and Janus-1.5B, indicating that the effectiveness of the proposed multi-modal multi-trajectory planning is not tied to a particular VLA backbone. Instead, the gains mainly come from its ability to provide a unified manner for VLA-based end-to-end driving. Specifically, our method leverages multi-modal inputs to support more reliable multi-trajectory prediction, improving the diversity and robustness of ego motions. These results collectively demonstrate that our method is a general and flexible end-to-end driving framework, with good transferability across different VLA foundations, planning formulations, and inference paradigms.

\textbf{Effectiveness of Driving Risk Cost Map.}~We conduct an ablation
study to assess the impact of the optimization objectives:
Prior Cost, Driving Risk Cost (DRCM \& SRCM), and
Smooth Cost, as summarized in Tab.~\ref{table_abl_opt}. Results reveal that removing any individual objective leads to a noticeable drop in overall PDMS performance. In particular,
disabling the Prior Cost reduces PDMS
from 92.2 to 88.0, indicating the importance of safety-aware priors for effective trajectory
planning. Similarly, removing the Driving Risk Cost results in both -7.1 \& -6.6 PDMS decrease, highlighting the benefit of 3D detection and BEV map segmentation for improved ego guidance quality. Lastly, disabling the Smooth Cost also causes
a performance drop of -0.7. These results confirm that our optimization algorithm can effectively exploit ego's neighboring 3D position and semantic information, leading to more reliable and stable trajectory refinement.

\begin{table}[t]
\scriptsize
\footnotesize
\setlength{\tabcolsep}{1.2pt}
\centering
\renewcommand\arraystretch{1.1}
\caption{ Ablation study of planning losses $\mathcal L_{plan}$ on
NAVSIM dataset.}
\label{table_abl_multi_traj}
\begin{tabular}{ccc|cccccc}
\toprule [0.8pt]
\multicolumn{3}{c|}{Planning Loss}       & \multicolumn{6}{c}{Evaluation Metrics}                    \\ 
$L_{match}$ & $L_{diversity}$ & $L_{cons.}$ & NC~$\uparrow$   & DAC~$\uparrow$  & TTC~$\uparrow$  & Comf.~$\uparrow$ & \multicolumn{1}{c}{EP~$\uparrow$}   & PDMS~$\uparrow$ \\ \midrule
\ding{51}        &              &       & 99.0 & 97.0 & 97.6 & 100   & \multicolumn{1}{c}{85.0} & 86.9  \\
\ding{51}        & \ding{51}    &     & 99.0 & 97.2 & 97.8 & 100   & \multicolumn{1}{c}{85.4} & 87.5  \\
\rowcolor{gray!10}\ding{51}        & \ding{51}            & \ding{51}              & 99.0 & 98.5 & 98.3 & 100   & \multicolumn{1}{c}{88.6} & 92.2  \\ \bottomrule [0.8pt]
\end{tabular}
\vspace{-0.3cm}
\end{table}

\begin{figure}[t]
    % \vspace{-0.9cm}
    \centering
    \includegraphics[width=0.49\textwidth]{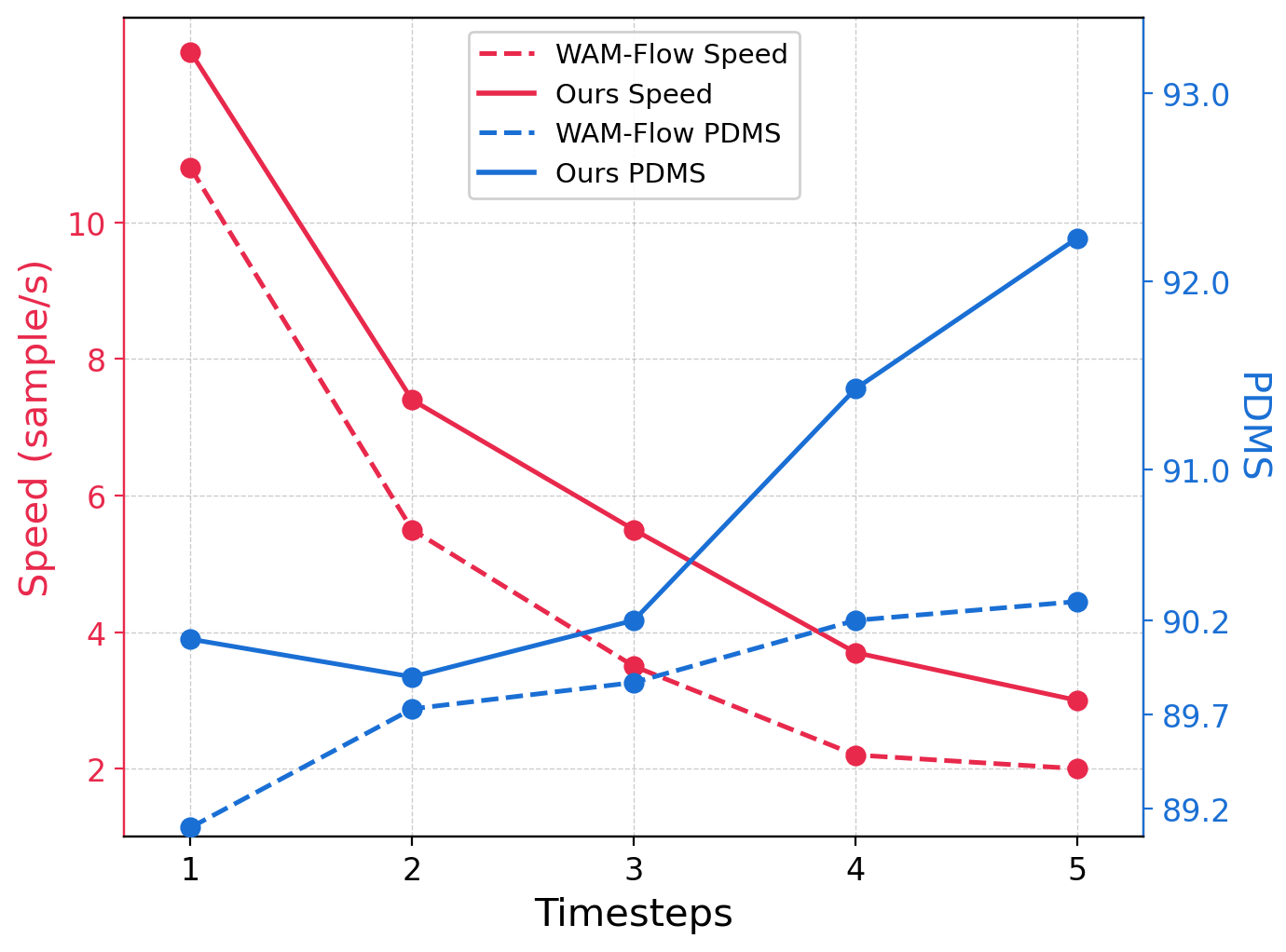} 
    \vspace{-0.6cm}
    \caption{Ablations of performance-speed trade-off on NAVSIM dataset.}
    \label{FIG_abl_speed}
    % \vspace{-0.4cm}
\end{figure}

\begin{table}[t]
\scriptsize
\footnotesize
\setlength{\tabcolsep}{3.2pt}
\centering
\renewcommand\arraystretch{1.1}
\caption{Ablation study of different VLA foundations on
NAVSIM dataset.}
\label{table_abl_vlm}
\begin{tabular}{c|cccccc}
\toprule [0.8pt]
\multirow{2}{*}{VLA Backbone} & \multicolumn{6}{c}{Evaluation Metrics} \\
                              & NC~$\uparrow$   & DAC~$\uparrow$  & TTC~$\uparrow$  & Comf.~$\uparrow$ & \multicolumn{1}{c}{EP~$\uparrow$}   & PDMS~$\uparrow$ \\ \midrule
Qwen2.5-3B                    & 98.9  & 98.4  & 98.0 & 99.9   & 88.5 & 91.8  \\
LLaVA-7B                      & 99.2  & 98.3  & 98.2 & 100   & 88.4 & 92.1  \\
\rowcolor{gray!10}Janus-1.5B  & 99.0 & 98.5 & 98.3 & 100   & 88.6 & 92.2  \\ \bottomrule [0.8pt]
\end{tabular}
\vspace{-0.2cm}
\end{table}

\begin{table}[!t]
\scriptsize
\footnotesize
\setlength{\tabcolsep}{1.2pt}
\centering
\renewcommand\arraystretch{1.1}
\caption{Ablation study of Perception-Oriented Trajectory Refinement on
NAVSIM dataset. "DRCM" and "SRCM" denote Detection Risk Cost Map \& Semantic Risk Cost Map, respectively.}
\label{table_abl_opt}
\begin{tabular}{cccc|cccccc}
\toprule [0.8pt]
\multicolumn{4}{c|}{Optimization Objectives} & \multicolumn{6}{c}{Evaluation Metrics} \\
Prior     & DRCM     & SRCM     & Smooth     & NC~$\uparrow$   & DAC~$\uparrow$  & TTC~$\uparrow$  & Comf.~$\uparrow$ & \multicolumn{1}{c}{EP~$\uparrow$}   & PDMS~$\uparrow$ \\ \midrule
\rowcolor{gray!10}\ding{51} & \ding{51}    & \ding{51}  & \ding{51}  & 99.0 & 98.5 & 98.3 & 100 & 88.6 & 92.2  \\ \midrule
\ding{55}         & \ding{51}        & \ding{51}        & \ding{51}  & 97.3  & 96.9  & 96.2 & 100 & 87.8 & 88.0  \\
\ding{51}         & \ding{55}        & \ding{51}        & \ding{51}  & 96.1  & 95.7  & 94.9 & 99.7 & 87.2 & 85.1  \\
\ding{51}         & \ding{51}        & \ding{55}        & \ding{51}  & 96.7  & 96.0  & 95.4 & 98.9 & 86.4 & 85.6  \\
\ding{51}         & \ding{51}        & \ding{51}        & \ding{55}  & 98.9  & 98.5  & 98.3 & 97.6 & 88.0 & 91.5  \\ \bottomrule [0.8pt]
\end{tabular}
\vspace{-0.15cm}
\end{table}

\textbf{Contribution of Proposed Trajectory Refinement.}~Fig.~\ref{FIG_abl_opt} demonstrates the contribution of our multi-trajectory optimization. Starting from either the trajectories obtained from our proposed multi-trajectory generation module (a)~\&~(b) or those produced by the original DFM (c)~\&~(d), the proposed optimization leverages neighboring spatial and semantic cues to refine the predicted motion. In both cases, the optimized trajectories exhibit significantly improved spatial consistency and better alignment with the ground-truth driving path. The optimization effectively suppresses implausible trajectory branches, corrects deviations caused by local prediction uncertainty, and guides the trajectories toward smoother and safer driving behaviors. Notably, the optimization consistently improves trajectory quality regardless of the initial trajectory generator, demonstrating its strong robustness and generalization ability across different planning frameworks.

\textbf{Performance-Efficiency Trade-Off.}~Fig.~\ref{FIG_abl_speed} presents a comparison with WAM-Flow~\cite{xu2026wam} under different denoising steps. Our method consistently exhibits a more favorable performance-speed trade-off. With 1-step denoising, our method achieves 90.1 PDMS at 11.5 samples/s. 
Increasing the number of refinement steps further improves the driving performance, with our method reaching 92.23 PDMS at 5 steps, while still maintaining a higher inference speed (3.0 v.s. 2.0 samples/s).
This trend indicates that our method can flexibly adjust the inference budget according to scenario complexity: few-step denoising provides efficient inference for straightforward scenarios, whereas additional refinement process offer substantial performance gains for more challenging driving situations. Overall, these results highlight the scalability and efficiency of our multi-tarjectory planning and optimizatiton.

\newpage

\begin{figure*}[ht]
    % \vspace{-0.9cm}
    \centering
    \includegraphics[width=\textwidth]{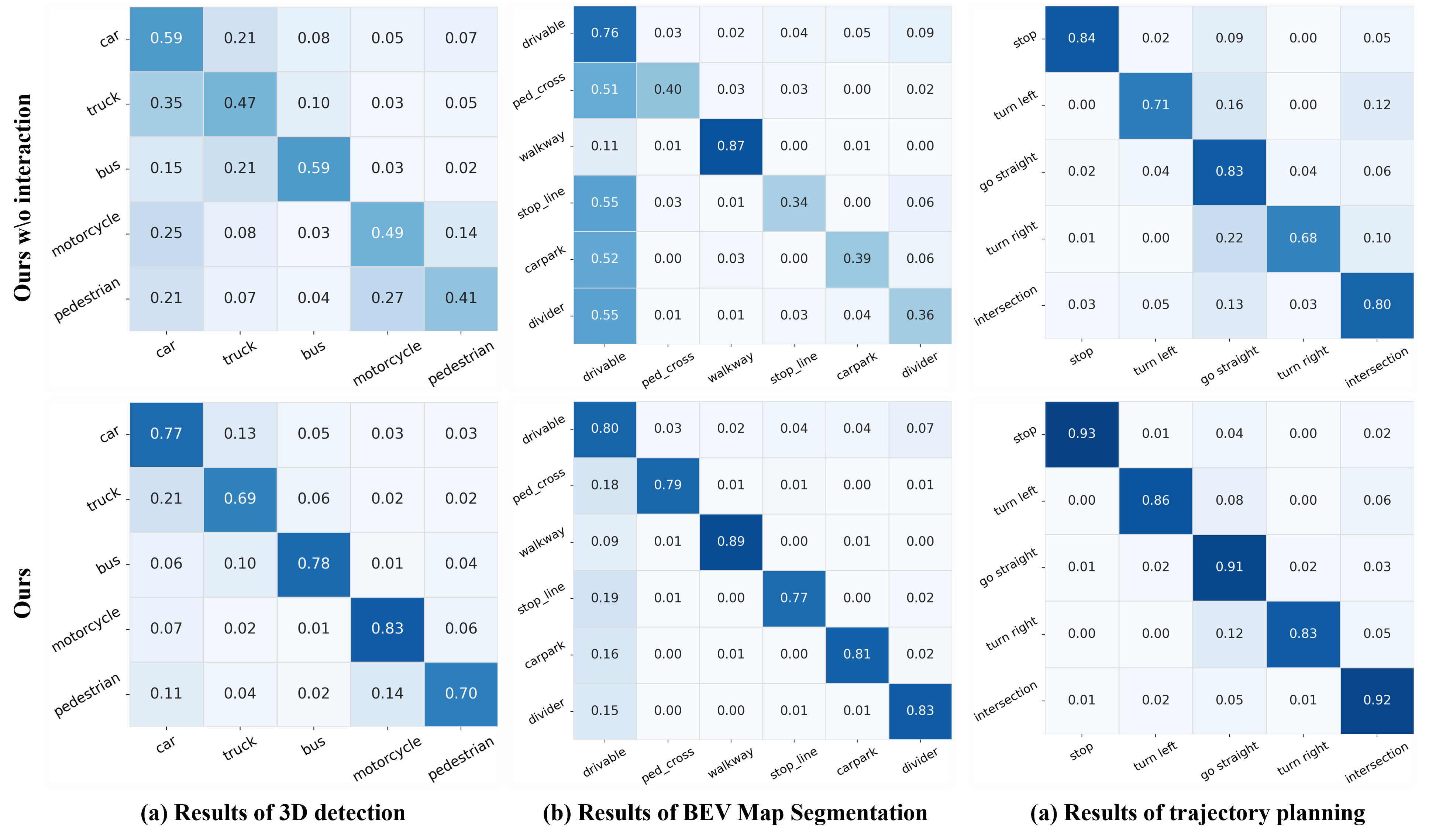} 
    \vspace{-0.6cm}
    \caption{Visualization of confusion matrices for evaluating our multi-modal interaction on the nuScenes dataset for 3D detection and BEV map segmentation, and on the NAVSIM test set for trajectory planning. We compare the full model with its variants without multi-modal interaction.}
\label{FIG_abl_confusion}
    % \vspace{-0.4cm}
\end{figure*}

\begin{table}[]
\scriptsize
\footnotesize
\setlength{\tabcolsep}{1.3pt}
\centering
\renewcommand\arraystretch{1.1}
\caption{Ablation study of different interaction manners on both
NAVSIM and nuScenes dataset. Where "DCMT" denotes Distribution-Consistent Modality Transfer.}
\label{table_abl_manner}
\begin{tabular}{c|ccc|cc}
\toprule [0.8pt]
\multirow{2}{*}{Interactive Manner} & \multicolumn{3}{c|}{nuScenes Dataset} & \multicolumn{2}{c}{NAVSIM  Dataset} \\
                                    & mAP~$\uparrow$        & NDS~$\uparrow$        & mIoU~$\uparrow$        & PDMS~$\uparrow$             & EPDMS~$\uparrow$            \\ \midrule
BEVFusion                           & 69.2       & 71.8       & 62.7        & 90.3             & 85.7              \\
LiDAR-only counterpart              & 70.7       & 75.9       & 68.0        & 91.0             & 86.4              \\
w/o DCMT                            & 73.6       & 77.7       & 71.0        & 92.0             & 86.9              \\
w/o prompts interaction             & 73.9       & 77.9       & 71.1        & 90.1             & 86.0              \\ \midrule
\rowcolor{gray!10}Ours              & 74.0       & 78.0       & 71.1        & 92.2            & 87.0              \\ \bottomrule [0.8pt]
\end{tabular}
\vspace{-0.3cm}
\end{table}

\subsubsection{Ablations of Multi-Modality Interaction}

\textbf{Influence of Different Interaction Manners.}~We investigate the influence of different modality interaction manners, with the results shown in Tab.~\ref{table_abl_manner}. Compared with the LiDAR-only counterpart and BEVFusion, our method achieves consistent improvements across both perception and planning metrics, demonstrating the benefit of multi-modal interaction. Removing DCMT results in a slight performance degradation, with mAP, NDS, and mIoU decreasing from 74.0, 78.0, and 71.1 to 73.6, 77.7, and 71.0, respectively, while PDMS and EPDMS decrease from 92.2 and 87.0 to 92.0 and 86.9. This demonstrates that distribution-consistent modality transfer further improves cross-modal feature alignment. Removing prompt interaction leads to a more noticeable degradation in planning, reducing PDMS from 92.2 to 90.1 and EPDMS from 87.0 to 86.0, highlighting the importance of semantic guidance for trajectory planning. 
%Overall, the complete model achieves the best performance across all evaluated metrics, validating the effectiveness of the proposed multi-modal interaction design.

\begin{table}[t]
\scriptsize
\footnotesize
\setlength{\tabcolsep}{1.3pt}
\centering
\renewcommand\arraystretch{1.1}
\caption{Ablation study of our multi-modality interaction on
NAVSIM and nuScenes. "AGOT" and "DCMT" denote Affinity-Guided Optimal Transport and Distribution-Consistent Modality Transfer.}
\label{table_abl_interaction}
\begin{tabular}{cc|cc|ccc|cc}
\toprule [0.8pt]
\multicolumn{2}{c|}{AGOT} & \multicolumn{2}{c|}{DCMT} & \multicolumn{3}{c|}{nuScenes Dataset} & \multicolumn{2}{c}{NAVSIM  Dataset} \\ 
${\mathcal L_{reg}}$         & ${\mathcal L_{ot}}$        & ${\mathcal L_{unf}}$      & ${\mathcal L_{consist}}$      & mAP~$\uparrow$        & NDS~$\uparrow$        & mIoU~$\uparrow$        & PDMS~$\uparrow$             & EPDMS~$\uparrow$            \\ \midrule
\ding{51}             & \ding{55}           & \ding{51}          & \ding{51}  & 72.0  & 75.4  & 68.6  & 91.0 &85.8              \\
\ding{55}            & \ding{51}           & \ding{51}          & \ding{51}   & 71.3  & 76.0  & 67.5  & 91.1 & 86.0              \\
\ding{51}             & \ding{51}           & \ding{51}          & \ding{55}  & 73.0  & 77.5  & 70.7  & 91.8 & 86.7              \\
\ding{51}             & \ding{51}           & \ding{55}         & \ding{51}   & 73.2  & 77.7  & 70.9  & 91.7 & 86.4              \\ \midrule
\rowcolor{gray!10}\ding{51}             & \ding{51}           & \ding{51}          & \ding{51}              & 74.0       & 78.0       & 71.1        & 92.2            & 87.0              \\ \bottomrule [0.8pt]
\end{tabular}
\vspace{-0.3cm}
\end{table}

\textbf{Impact of the Affinity-Guided Optimal Transport.}~We evaluate the contributions of the regularization loss $\mathcal L_{reg}$ and OT loss $\mathcal L_{ot}$ in Affinity-Guided Optimal Transport, as shown in Tab.~\ref{table_abl_interaction}. Removing either component leads to performance degradation across both perception and planning metrics, demonstrating the complementary roles of the two objectives. Specifically, removing $\mathcal L_{reg}$ reduces the mAP, NDS, and mIoU by 2.0, 2.6, and 2.5, respectively, while decreasing PDMS and EPDMS by 1.1 and 1.2. Removing $\mathcal L_{ot}$ causes a slightly larger degradation, with mAP, NDS, and mIoU dropping by 2.7, 2.0, and 3.6, respectively, and PDMS decreasing by 1.2. The larger impact of $\mathcal L_{ot}$ on overall performance indicates that explicit optimal transport supervision is particularly important for establishing accurate cross-modal correspondences, while $\mathcal L_{reg}$ further stabilizes the affinity learning. These results validate the complementary roles of the two losses in AGOT.

\textbf{Effectiveness of the Distribution-Consistent Modality Transfer.}~We further investigate the two components of Distribution-Consistent Modality Transfer, as shown in Tab.~\ref{table_abl_interaction}. Removing either component consistently degrades performance across both perception and planning metrics, demonstrating the complementary contributions of the two objectives. Specifically, removing $\mathcal L_{unf}$ decreases mAP, NDS, and mIoU by 0.8, 0.3, and 0.2, respectively, while reducing PDMS and EPDMS by 0.5 and 0.6. In comparison, removing $\mathcal L_{consist}$ results in slightly larger perception degradation, with mAP, NDS, and mIoU decreasing by 1.0, 0.5, and 0.4, while PDMS and EPDMS decrease by 0.4 and 0.3, respectively. These results suggest that $\mathcal L_{unf}$ and $\mathcal L_{consist}$ play a slightly more important role in improving planning performance by aligning heterogeneous modalities toward a unified latent distribution for cross-modal representations.
%These results confirm that unified distribution alignment and consistency regularization are both essential for reliable cross-modal interaction.

\textbf{Visualization analysis of Confusion Matrices.}~To further validate the effectiveness of our multi-modality interaction, we visualize the confusion matrices of 3D detection, BEV map segmentation, and trajectory planning for the model without interaction and our full model. As shown in Fig.~\ref{FIG_abl_confusion}, our method exhibits stronger diagonal responses and fewer off-diagonal confusions across the three tasks. This improvement indicates that the proposed interaction mechanism effectively exploits complementary information from heterogeneous modalities, leading to more discriminative object representations and more accurate BEV semantic understanding. Consequently, the improved multi-modal representations provide more reliable semantic and spatial cues for downstream trajectory planning.

\section{Limitations and Further Work}

Despite its effectiveness, our framework has several limitations. First, the optimal transport-based interaction introduces additional computational overhead, particularly for dense multimodal tokens. Future work will explore more efficient sparse or hierarchical transport strategies. Second, the current trajectory refinement relies on perception-derived risk information and may be sensitive to perception errors. Future work will investigate joint uncertainty estimation and dynamic interaction modeling to further improve the robustness and safety of VLA-based autonomous driving.

\section{Conclusion}

In this work, we presented an interpretable VLA-based end-to-end autonomous driving framework for reliable multimodal perception, reasoning, and planning. We introduced Affinity-Guided Optimal Transport to establish structured two-way interactions between heterogeneous modalities and proposed Distribution-Consistent Modality Transfer to align multimodal features in a unified Gaussian latent space. Besides, we also developed the Multi Modality Multi Trajectory Planning with Perception-Oriented Trajectory Refinement to achieve trajectory optimization. Extensive experiments demonstrate that the proposed framework improves multimodal interaction and driving decision reliability to challenging long-tail scenarios while maintaining competitive inference efficiency. 
%These results highlight the potential of structured multimodal interaction and explicit trajectory refinement for developing more reliable and interpretable VLA-based autonomous driving systems.

% % use section* for acknowledgment
% \ifCLASSOPTIONcompsoc
%   % The Computer Society usually uses the plural form
%   \section*{Acknowledgments}
% \else
%   % regular IEEE prefers the singular form
% \fi

% % Can use something like this to put references on a page
% % by themselves when using endfloat and the captionsoff option.
% \ifCLASSOPTIONcaptionsoff
%   \newpage
% \fi

\bibliographystyle{Bibliography/IEEEtran}
\bibliography{Bibliography/IEEEabrv,Bibliography/bare_jrnl_compsoc}\ 

\vspace{-0.8cm}
\begin{IEEEbiography}[{\includegraphics[width=1in,height=1.25in,clip,keepaspectratio]{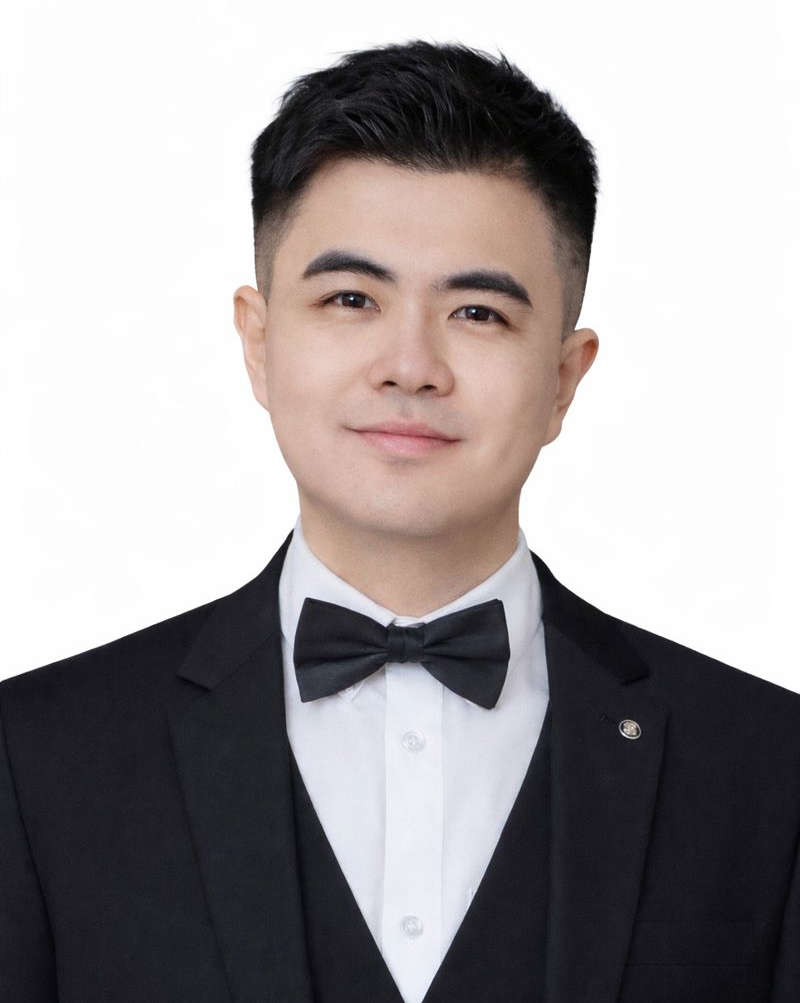}}]
{Jingtao Sun} received the B.S., M.S. and Ph.D degree in the National Engineering Research Center for Robot Visual Perception and Control from Hunan University, Changsha, China. He is currently a research fellow at the Department of Electrical and Computer Engineering (ECE), National University of Singapore (NUS). His research interests include 3D computer vision, robotics and multi-modal.
\end{IEEEbiography}
\vspace{-0.8cm}
\begin{IEEEbiography}[{\includegraphics[width=1in,height=1.25in,clip,keepaspectratio]{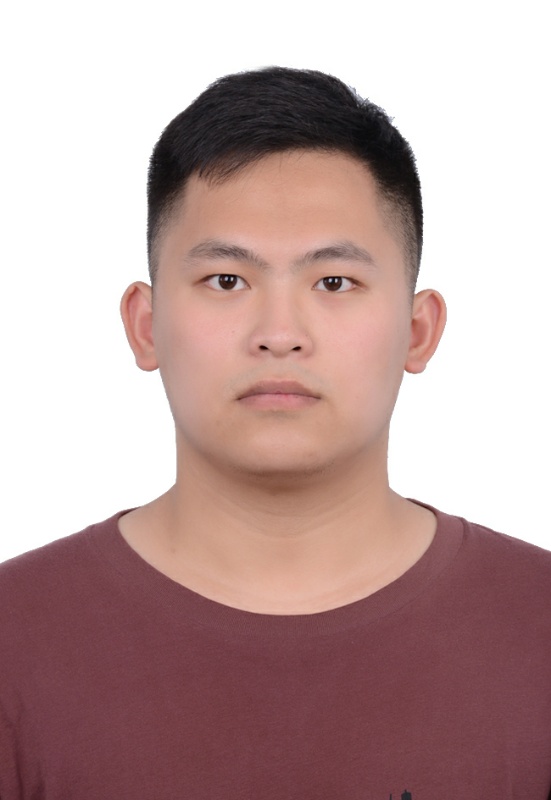}}]
{Xiaohai He} received the B.S. degree from Jiangsu University of Science and Technology, China, and the M.S. degree from Syracuse University, USA. He is currently pursuing his Ph.D. degree in School of Mechatronical Engineering, Beijing Institute of Technology, Beijing, China. He also is a visiting Ph.D student at the Department of Electrical and Computer Engineering (ECE), National
University of Singapore (NUS). His research interests focus on autonomous driving and robot navigation.
\end{IEEEbiography}
\vspace{-0.8cm}
\begin{IEEEbiography}[{\includegraphics[width=1in,height=1.25in,clip,keepaspectratio]{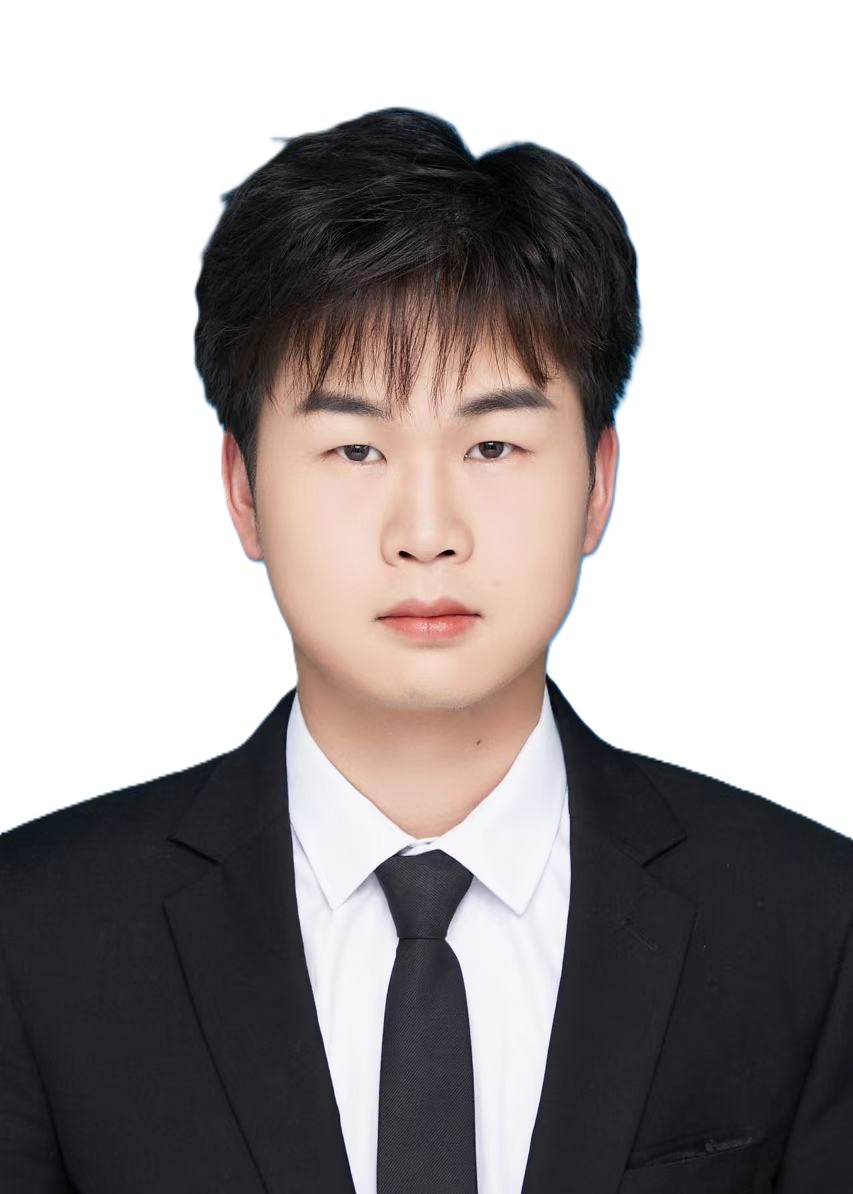}}]
{Yike Zhang} received the B.S. and M.S. degree in Control Science and Engineering from the School of Automation, Huazhong University of Science and Technology, Wuhan, China. He is currently completing his Ph.D. at the National Engineering Research Center for Robot Vision and Control, Hunan University, Changsha, China. He also is a visiting Ph.D student at the Department of Electrical and Computer Engineering (ECE), National
University of Singapore (NUS). His research interests focus on 3D computer vision and robot learning.
\end{IEEEbiography}
\vspace{-0.8cm}

\begin{IEEEbiography}[{\includegraphics[width=1in,height=1.25in,clip,keepaspectratio]{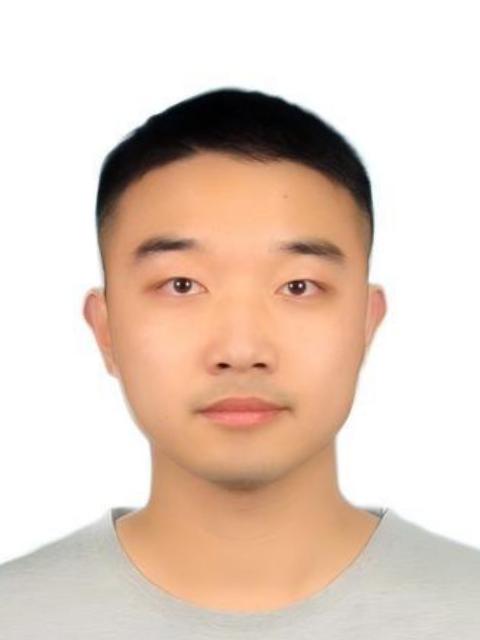}}]{Dong Huang} received the B.S. degree in Information Engineering from the School of Electronic Information and Electrical Engineering, Shanghai Jiao Tong University, Shanghai, China, and the M.S. degree in Computer Engineering from the Department of Electrical and Computer Engineering, National University of Singapore, Singapore. He is currently pursuing the Ph.D. degree in the Department of Electrical and Computer Engineering, National University of Singapore, Singapore. His research interests focus on the safety and security of multimodal agents.
\end{IEEEbiography}
\vspace{-0.8cm}
\begin{IEEEbiography}[{\includegraphics[width=1in,height=1.25in,clip,keepaspectratio]{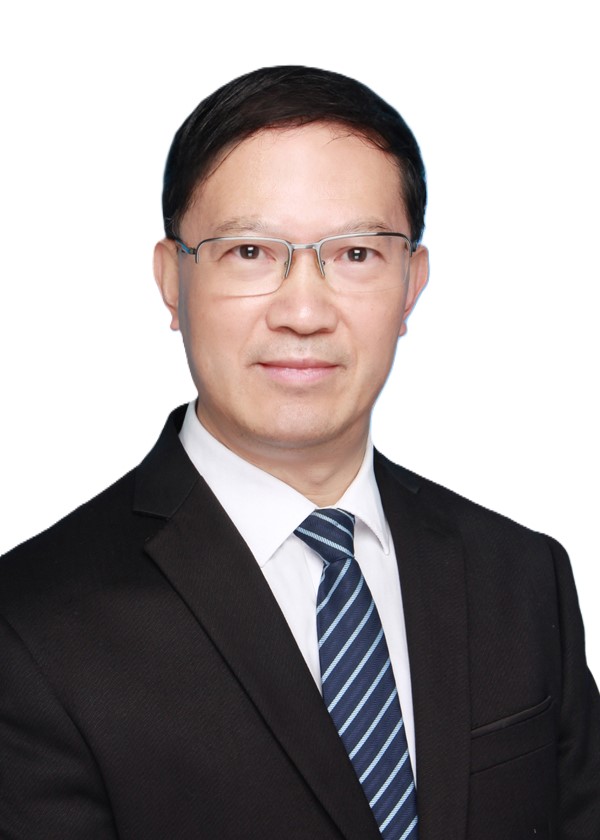}}]
{Yaonan Wang} received the Ph.D. degree in electrical engineering from Hunan University, Changsha, China, in 1994. He was a PostDoctoral Research Fellow with the Normal University of Defence Technology, Changsha, from 1994 to 1995. From 1998 to 2000, he was a Senior Humboldt Fellow in Germany, and, from 2001 to 2004, he was a Visiting Professor with the University of Bremen, Bremen, Germany. Since 1995, he has been a Professor with the College of Electrical and Information Engineering, Hunan University. He is an Academician with the Chinese Academy of Engineering. His current research interests include robotics and computer vision.
\end{IEEEbiography}
\vspace{-0.8cm}

\begin{IEEEbiography}[{\includegraphics[width=1in,height=1.25in,clip,keepaspectratio]{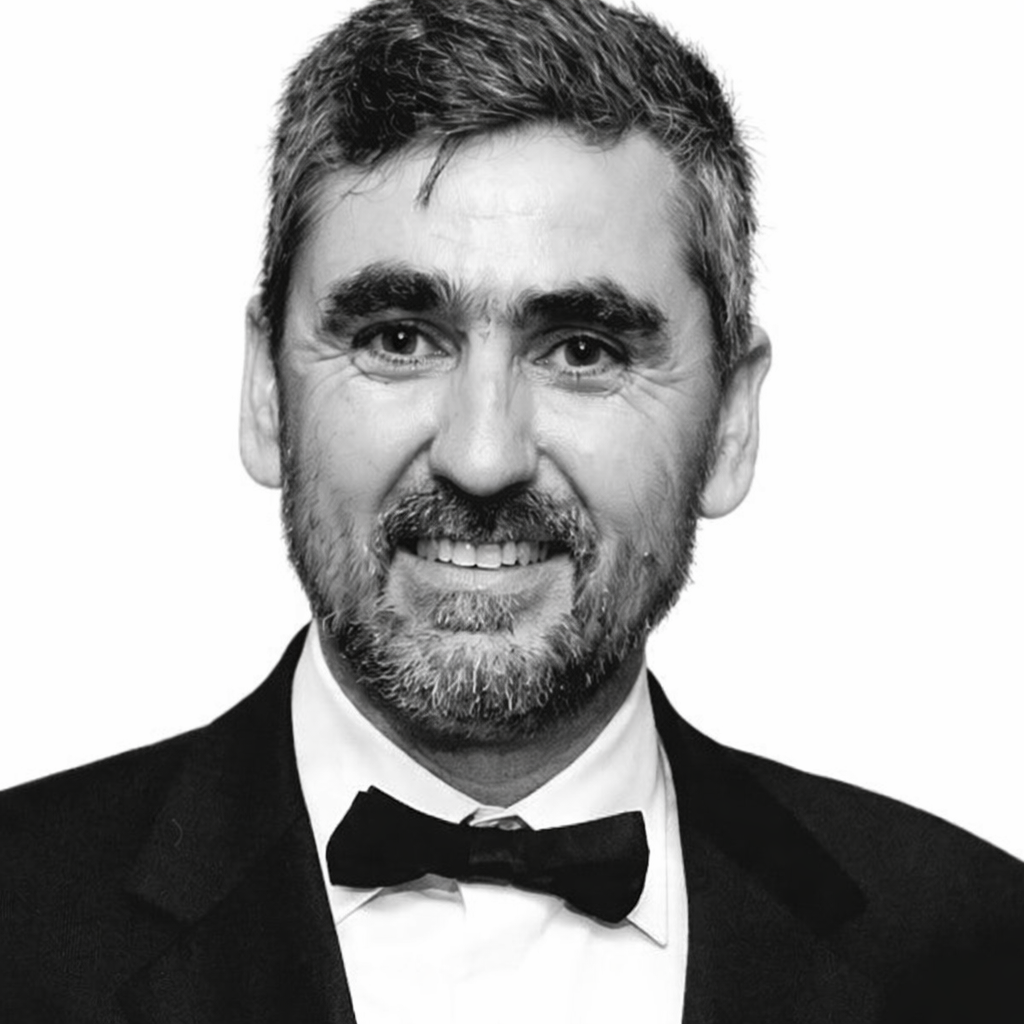}}]
{Ajmal Mian} is a Professor of Computer Science at The University of Western Australia. He currently serve as an Associate Editor of IEEE TPAMI. He also serves as an Senior Editor of IEEE TNNLS, Associate Editor of IEEE TIP and the Pattern Recognition journal. He serves as Area Chair for CVPR, ECCV, ACM MM, etc.
His research interests include computer vision, machine learning, 3D shape analysis, human action recognition, video description and hyperspectral image analysis. 
\end{IEEEbiography}

\vspace{-0.8cm}
\begin{IEEEbiography}[{\includegraphics[width=1in,height=1.25in,clip,keepaspectratio]{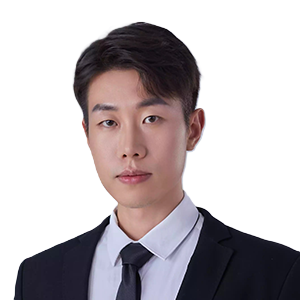}}]
{Mike Zheng Shou} reveived the Ph.D degree in Electrical Engineering from Columbia University, New York City and prior to joining NUS, he was a Research Scientist at Facebook AI, Menlo Park, California.He is currently a tenure-track Assistant Professor with the Department of Electrical and Computer Engineering (ECE) at National University of Singapore (NUS). He serve as an associate editor of IEEE TPAMI. He also serves as Area Chair for CVPR, ECCV, ICCV, ACM MM, etc. His research interests include computer vision and multi-modal.
\end{IEEEbiography}


%\vfill


\end{document}